\documentclass[10pt,twocolumn,letterpaper]{article}

\usepackage{iccv}
\usepackage{times}
\usepackage{epsfig}
\usepackage{graphicx}
\usepackage{amsmath}
\usepackage{amssymb}
\usepackage{booktabs}
\usepackage{multirow}
\usepackage{csquotes}
\usepackage{threeparttable}
\usepackage[normalem]{ulem}
\usepackage{caption}
\usepackage{enumerate}
\usepackage[export]{adjustbox}
\usepackage{paralist}
\usepackage[accsupp]{axessibility}
\usepackage{cite} 

\usepackage[pagebackref=true,breaklinks=true,letterpaper=true,colorlinks,bookmarks=false]{hyperref}

\iccvfinalcopy 

\def\iccvPaperID{2389} 

\ificcvfinal\fi

\begin{document}

\title{RPEFlow: Multimodal Fusion of RGB-PointCloud-Event \\
for Joint Optical Flow and Scene Flow Estimation}

\author{Zhexiong Wan\textsuperscript{1} \space\space\space\space Yuxin Mao\textsuperscript{1} \space\space\space\space Jing Zhang\textsuperscript{2} \space\space\space\space Yuchao Dai\textsuperscript{1}$^{\dagger}$ \\
\textsuperscript{1}Northwestern Polytechnical University \& Shaanxi Key Laboratory of \\  Information Acquisition and Processing  \space\space\space\space \textsuperscript{2}Australian National University
}

\maketitle
\ificcvfinal\thispagestyle{empty}\fi

\noindent\let\thefootnote\relax\footnotetext{{${\dagger}$ Corresponding author \tt(daiyuchao@gmail.com).}}

\begin{abstract}
Recently, the RGB images and point clouds fusion methods have been proposed to jointly estimate 2D optical flow and 3D scene flow.
However, as both conventional RGB cameras and LiDAR sensors adopt a frame-based data acquisition mechanism, their performance is limited by the fixed low sampling rates, especially in highly-dynamic scenes. 
By contrast, the event camera can asynchronously capture the intensity changes with a very high temporal resolution, providing complementary dynamic information of the observed scenes.
In this paper, we incorporate \textbf{R}GB images, \textbf{P}oint clouds and \textbf{E}vents for joint optical flow and scene flow estimation with our proposed multi-stage multimodal fusion model, \textbf{RPEFlow}. 
First, we present an attention fusion module with a cross-attention mechanism to implicitly explore the internal cross-modal correlation for 2D and 3D branches, respectively. 
Second, we introduce a mutual information regularization term to explicitly model the complementary information of three modalities for effective multimodal feature learning. 
We also contribute a new synthetic dataset to advocate further research. 
Experiments on both synthetic and real datasets show that our model outperforms the existing state-of-the-art by a wide margin. 
Code and dataset is available at \url{https://npucvr.github.io/RPEFlow}.

\end{abstract}

\section{Introduction}
\label{sec:intro}

Optical flow estimation, \ie, estimating the dense 2D motion between consecutive image frames, has been extensively studied and significantly advanced with the development of deep neural networks~\cite{flow:Sun_PWCNet_CVPR_2018, flow:Teed_RAFT_ECCV_2020, flow:huang_flowformer_ECCV_2022}. 
Scene flow estimation, on the other hand, aims to estimate the 3D motion field with various input configurations, ranging from monocular images~\cite{scene:hur_selfscene_CVPR_2020, scene:guizilini_DRAFT_RAL_2022}, stereo images~\cite{scene:ilg_occlusions_ECCV_2018, scene:ma_deeprigid_CVPR_2019}, two frames of point clouds~\cite{scene:wu_pointpwc_eccv_2020, scene:li_sctn_AAAI_2022}, images combined with depth maps~\cite{scene:teed_raft3d_cvpr_2021, scene:yang_flowexpan_CVPR_2020} or point clouds~\cite{scene:rishav_deeplidarflow_iros_2020, scene:liu_camliflow_cvpr_2022}. 
Both are fundamental to downstream applications such as autonomous driving~\cite{flowdatasets:Geiger_kitti2012_cvpr_2012, flowdatasets:menze_object_kitti_2015}, object tracking~\cite{porzi_trackseg_auto_CVPR_2020, wang_multiple_correlation_CVPR_2021}, scene reconstruction~\cite{zhang_flowfusion_ICRA_2020, li_neuralscenefields_CVPR_2021}, \etc. 

Due to the strong correlation between 2D and 3D motion, \ie, 2D motion can be regarded as the projection of 3D motion on the image plane, recent works~\cite{scene:rishav_deeplidarflow_iros_2020, scene:teed_raft3d_cvpr_2021, scene:liu_camliflow_cvpr_2022} make efforts to jointly estimate optical flow and scene flow by combining RGB images and point clouds (or depth maps). 
Their success indicates that joint 2D and 3D motion estimation within a framework can obtain more accurate results than separate tasks. 
However, as both conventional RGB cameras and LiDAR (or depth) sensors adopt a fixed frame-by-frame data acquisition mechanism, these methods show unsatisfactory performance when dealing with complex motion scenes (see Fig.~\ref{viz:things_kubric}), which motivates us to alleviate this problem by introducing the event camera. 

Event camera, as a bio-inspired imaging sensor, can asynchronously capture the brightness change with very high temporal resolution (in the order of $\mu s$) and output an event signal quickly~\cite{gallego_eventsurvey_TPAMI_2022}. 
As each pixel adapts its sampling rate according to the captured changes, the amount of output events usually depends on the complexity of motion (the faster the motion, the more triggered events), thus providing abundant motion information of the observed scene. 
Based on this, some works use event data alone to estimate optical flow~\cite{eventflow:Zhu_EVFlowNet_CVPR_2019, eventflow:Gehrig_DenseRAFTFlow_3DV_2021}, but they show limitations in estimating reliable motion at regions with no events~\cite{eventflow:liu_adaptiveblock_BMVC_2018}.
As compensation for this, image and event data are fused together to estimate dense optical flow~\cite{eventflow:Pan_SingleImageFlow_CVPR_2020, eventflow:Wan_DCEIFlow_TIP_2022}. 
As far as we know, there is no method to incorporate event data within a multimodal learning framework for both 2D and 3D motion estimation.

In this paper, we propose to fuse RGB images, point clouds and events for joint optical flow and scene flow estimation. 
We find the ability of the event camera to asynchronously capture the brightness changes caused by motion makes it complementary to image cameras and LiDAR sensors, especially for complex dynamics and high-contrast brightness changes. 
We believe that combining these three modalities together for 2D and 3D motion estimation meets the practical needs, which has been further confirmed by existing datasets, such as MVSEC~\cite{eventdatasets:Zhu_MVSEC_RAL_2018} and DSEC~\cite{eventflow:Gehrig_DenseRAFTFlow_3DV_2021} that contain these data for driving scenarios. 

We formulate this task as a representation-based multimodal learning problem, and exploit the complementary information between these three very different modalities implicitly and explicitly.
We aim to exploit the relationships between multimodal and multi-dimensional space observations (images and events in 2D with point clouds in 3D) and explore their contributions to 2D and 3D motion. 
Specifically, in our RPEFlow framework, we first propose a multimodal attention fusion module with \textit{cross-attention mechanism} to implicitly explore the correlations between three modalities, based on which a pyramid multi-stage fusion structure is introduced to extensively modeling.
We observe that each modality can contribute a part to 2D and 3D motion estimation, making representation learning~\cite{Isolating_Sources} suitable for our multimodal learning framework.
Then we introduce cross-modal mutual information minimization in feature space to explicitly maximize the complementary information.
We also contribute a new synthetic dataset with simulations that conform to the gravity model and collision detection and contain a larger variety of moving objects and richer annotations than FlyingThings3D~\cite{flowdatasets:Mayer_FlyingThing3d_CVPR_2016}.
Extensive experimental results validate both our implicit multimodal attention fusion and explicit representation regularization towards effective multimodal learning, leading to a new benchmark on both synthetic and real-captured datasets. 

Our main contributions are summarized as follows:
\begin{compactenum}[1)]
    \item We propose to incorporate event cameras with RGB cameras and LiDAR sensors to jointly estimate optical flow and scene flow for complex dynamic scenes, which constitutes a new and practical problem. 
    \item An implicit multimodal attention fusion module and an explicit representation learning via mutual information regularization are presented in our RPEFlow model, achieving extensive cross-modal relationship modeling.
    \item We contribute a large-scale synthetic dataset with ground-truth motion annotations. Experimental results on both synthetic and real datasets show that the proposed RPEFlow outperforms existing state-of-the-art and demonstrates the effectiveness of event data for motion estimation of complex dynamics.
\end{compactenum}

\section{Related Work}

\subsection{Unimodal 2D/3D Motion Estimation}

\noindent \textbf{Image only.}
For learning-based 2D optical flow estimation, FlowNet series~\cite{flow:dosovitskiy_flownet_iccv_2015, flow:Ilg_Flownet2_cvpr_2017} first propose end-to-end CNN models for regression. 
PWC-Net~\cite{flow:Sun_PWCNet_CVPR_2018} work on constructing feature pyramids with coarse-to-fine refinement.
RAFT~\cite{flow:Teed_RAFT_ECCV_2020} and its variants~\cite{flow:xu_gmflow_CVPR_2022, flow:huang_flowformer_ECCV_2022} build all-pairs correlation and update the optical flow iteratively.
For 3D scene flow estimation, learning-based studies~\cite{scene:ma_deeprigid_CVPR_2019, scene:ilg_occlusions_ECCV_2018} use a sequence of stereo images as input~\cite{scene:vedula_threed_TPAMI_2005}, achieving faster and better performance compared with earlier optimization-based methods~\cite{scene:huguet_variational_ICCV_2007, flowdatasets:menze_object_kitti_2015}. 
Some recent works~\cite{scene:brickwedde_monosf_ICCV_2019, scene:hur_selfscene_CVPR_2020, scene:guizilini_DRAFT_RAL_2022} use only monocular image sequences, which are more difficult to model accurate 3D structure and motion than stereo images. 
Due to the limited frame rate of input images and the difficulty in obtaining 3D structure, the performance of image-only methods is still unsatisfactory when dealing with complex dynamics. 

\noindent \textbf{Point Cloud only.}
Point cloud data from the LiDAR sensor is favorable for 3D motion estimation.
Due to the unique data structure, existing methods~\cite{scene:wu_pointpwc_eccv_2020, scene:puy_flot_eccv_2020, scene:li_sctn_AAAI_2022} focus on studying the model structure to represent the point cloud data for scene flow estimation. 
However, the point cloud lacks semantic information and leads to the difficulty of estimating accurate motion only by the structural information~\cite{scene:liu_camliflow_cvpr_2022}. 

\noindent \textbf{Event only.}
Event-based motion estimation is dedicated to extracting motion information from event-encoded brightness changes. 
Some early optimization-based methods estimate the motion flow of moving boundaries~\cite{eventflow:Gallego_Unifyingcontrastmax_cvpr_2018, eventflow:ieng_event3dflow4dsubspace_2017}.
Recent learning-based methods~\cite{eventflow:Zhu_EVFlowNet_CVPR_2019, eventflow:Lee_SpikeFlowNet_ECCV_2020, eventflow:Gehrig_DenseRAFTFlow_3DV_2021} are proposed to regress the dense optical flow directly. 
Even though the predictions of some of them are densely supervised, the sparse input event data leads to unreliable optical flow estimation in the regions without triggered events~\cite{eventflow:liu_adaptiveblock_BMVC_2018, eventflow:Pan_SingleImageFlow_CVPR_2020}. 

\begin{figure*}[t!]
\centering
\includegraphics[width=0.98\linewidth]{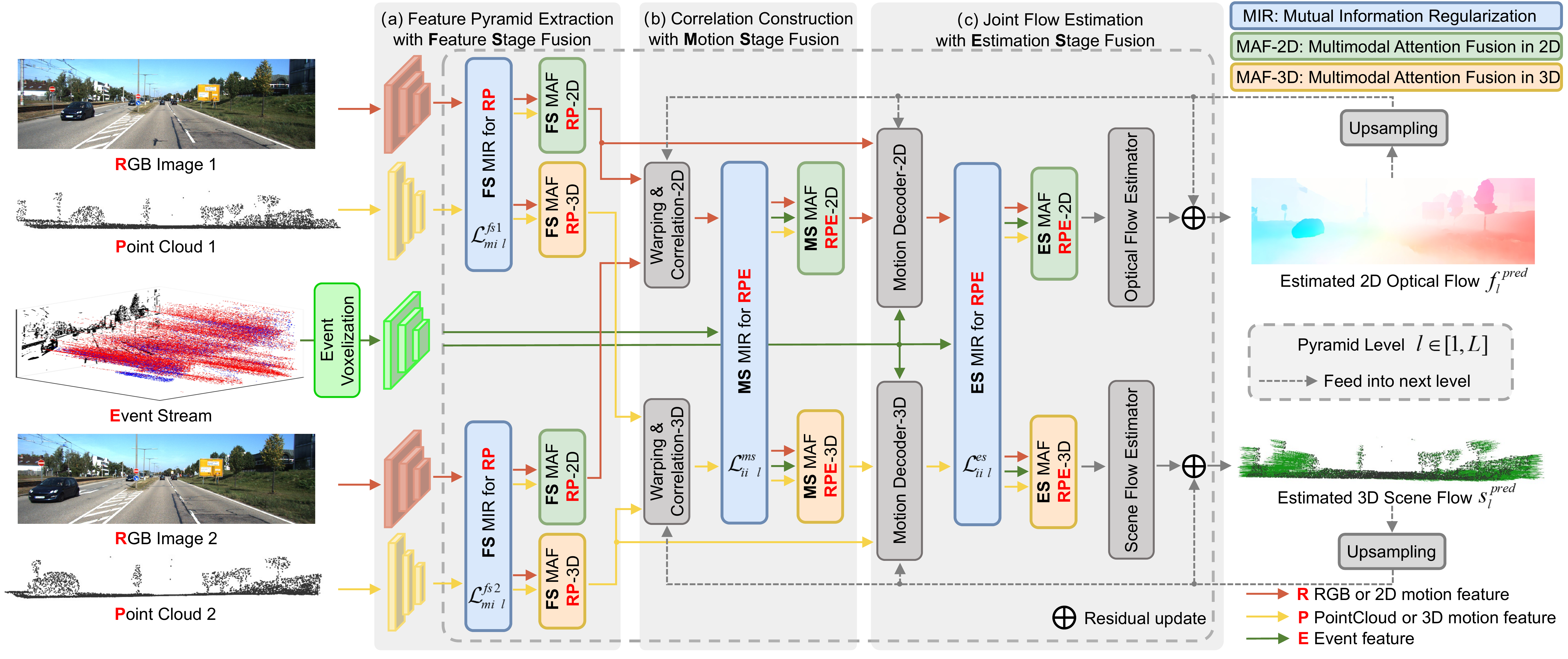}
\vspace{-5pt}
\caption{\textbf{Our RPEFlow Structure.} 
We learn motion correlations from the input three modalities (RGB-PointCloud-Event, RPE) by multi-stage fusion (FS, MS and ES) incorporating implicit multimodal attentional fusion (MAF) and explicit representing learning with mutual information regularization (MIR), and perform pyramidal updates from coarse to fine to estimate the optical flow in the 2D branch (top) and the scene flow in the 3D branch (bottom), respectively. 
Best viewed on screen. 
}
\vspace{-5pt}
\label{fig:network}
\end{figure*}

\subsection{Multimodal 2D/3D Motion Estimation}
As unimodal data only provides partial information, multimodal methods are presented to comprehensively learn from multiple observations. 
For optical flow estimation, event-based studies~\cite{eventflow:Bardow_simultaneousflowintensity_2016, eventflow:Pan_SingleImageFlow_CVPR_2020, eventflow:Wan_DCEIFlow_TIP_2022} combine the advantages of the image in dense representation and events in motion perception for reliable estimation. 
Besides, \cite{flow:poggi_sensorflow_ICCV_2021, flow:conti_sensorguide_Lidar_IROS_2022} incorporate gyroscope or depth sensor to guide the optical flow. 
For scene flow estimation, using a sequences of RGB-Depth images~\cite{scene:teed_raft3d_cvpr_2021, scene:yang_flowexpan_CVPR_2020} becomes another trend, then ~\cite{scene:rishav_deeplidarflow_iros_2020, scene:liu_camliflow_cvpr_2022} replace depth with point clouds to deal with the limited ideal range of depth camera when applied outdoors. 
But again, they are still limited by the frame-by-frame acquisition mechanism, which leads us to introduce event data. 

\subsection{Multimodal Fusion}
Given multimodal data, effective multimodal fusion is critical to extensively explore the contribution of each modality~\cite{multimodal_survey_2022}. Two main directions have been explored:
1) attention based~\cite{hori2017attention, wei2020multi, eventapp:sun_eventdeblurattention_eccv_2022} and 2) representation learning~\cite{representation_learning_yoshua, hu2017learning, mine_mutual_information, fusion:zhang_rgbdsali_ICCV_2021} based. 
For the former, a specific attention module~\cite{vaswani_attention_NeurIPS_2017, dosovitskiy_imagetransformer_iclr_2020} is designed to implicitly control the contribution of each modal.
For the latter, representation similarity is measured to explicitly constrain the reliability of the feature embedding.
Within this direction, mutual information (MI) estimation and optimization~\cite{hjelm2018learning, ba_mmm_nips03, cheng2020club, mine_mutual_information} is the widely studied strategy, which is typically used as a regularizer to encourage (via MI maximization) or limit dependency (via MI minimization) between variables. 

\section{Our Method}

We introduce two main strategies to achieve effective multimodal learning, one is implicit  multimodal attention fusion (Sec.~\ref{sec:attention}) and the other is explicit mutual information regularization (Sec.~\ref{sec:mutual_info_reg}). 
Based on them, we propose a pyramid multi-stage framework (see Fig.~\ref{fig:network}) for RGB-PointCloud-Event fusion and joint optical flow and scene flow estimation in 2D and 3D branches (Sec.~\ref{sec:joint_estimation}).

\subsection{Multimodal Attention Fusion (MAF)}
\label{sec:attention}

\begin{figure}[tbp]
\centering

\includegraphics[width=0.9\linewidth]{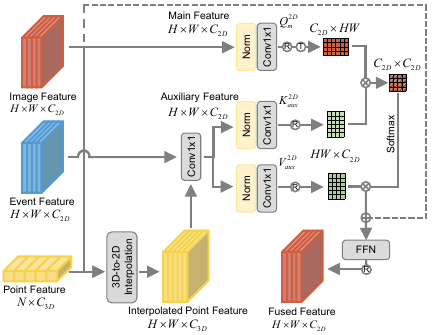}\\
\vspace{5pt}
\includegraphics[width=\linewidth]{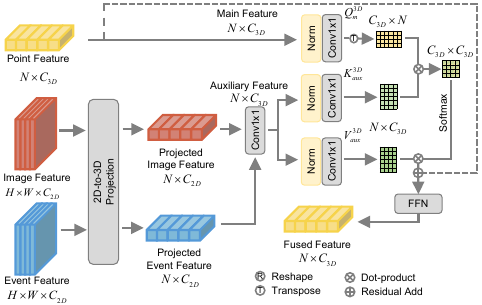}

\vspace{-5pt}
\caption{\textbf{Our proposed Multimodal Attention Fusion Module fuses both three modal features in 2D (top) and 3D (bottom) branches}, including feature projection and cross-attention fusion. 
In particular, only image and point features are fused in Feature Stage (FS).}
\label{fig:fusion}
\end{figure}

As shown in Fig.~\ref{fig:network}, we have both 2D and 3D branches for optical flow and scene flow estimation.
Due to the different data structures of the two branches~\cite{scene:wu_pointpwc_eccv_2020, scene:liu_camliflow_cvpr_2022}, we design symmetric attention fusion strategy for both two branches in Fig.~\ref{fig:fusion}, which consists of two steps, namely feature projection and cross-attention fusion.
In the 2D branch, we treat the RGB image as the primary modality and project point cloud feature into the image plane, then fuse with auxiliary features, \ie, event and point cloud.
For the 3D branch, we define the point cloud data as the primary modality, then project the others into the 3D space and fuse them together.

\noindent\textbf{Multimodal Attention Fusion in 3D branch (MAF RPE-3D).}
We take the fusion module in the 3D branch of a single pyramid level as a detailed example, where the encoded image feature is $e_r \in \mathbb{R}^{H \times W \times C_{2D}}$, event feature is $e_{ev} \in \mathbb{R}^{H \times W \times C_{2D}}$, and point cloud feature is $e_{pc} \in \mathbb{R}^{N \times C_{3D}}$ with the point positions $\mathbf{p} = \left\{\mathbf{p}_{x_i}, \mathbf{p}_{y_i}, \mathbf{p}_{z_i}\right\}^N \in \mathbb{R}^{N \times 3}$ in 3D space. 
Note that $H, W$ and $N$ are the feature size at the current pyramid level, not the original input size. 
We first use the point position to sample the corresponding image and event feature into 3D space with focal length $f$ and denote the projected point position at the image plane as:
\begin{equation}
\label{twod_projection_to_3d}
    \{(u_i, v_i)\}^N = \{ (f \frac{\mathbf{p}_{x_i}}{\mathbf{p}_{z_i}}, f \frac{\mathbf{p}_{y_i}}{\mathbf{p}_{z_i}} ) \} ^N \in \mathbb{R}^{N \times 2}.
\end{equation}

Thus, the projected image and event features are:
\begin{equation}
\begin{aligned}
\! e_{r}^{pj} \! = \! \{e_{r}(u_i, \! v_i)\}^N, e_{ev}^{pj} \! = \! \{e_{ev}(u_i, v_i)\}^N \! \in \! \mathbb{R}^{N \times C_{2D}},
\end{aligned}
\end{equation}
where $e_{r}(u_i, v_i)$ represents the feature obtained by bilinear interpolation sampling at $(u_i, v_i)$ position in image plane.

After projection, we feed the features of auxiliary modalities $e_{r}^{pj}$ and $e_{ev}^{pj}$ with the primary modality feature $X_{pri}^{3D} = e_{pc}$ into the attention fusion module. 
In the original self-attention mechanism~\cite{vaswani_attention_NeurIPS_2017, dosovitskiy_imagetransformer_iclr_2020}, all of the keys $K$, values $V$ and queries $Q$ come from the same modality.
Here we adapt it to accommodate inputs from multiple modalities and propose our cross-attention fusion structure.
Specifically, we first combine the auxiliary features and align the number of channels with the master feature by $1 \times 1$ convolution, yielding the aligned auxiliary feature $Y_{aux}^{3D} = W_{a} \left[e_{r}^{pj}, e_{ev}^{pj}\right] $ ($\mathbb{R}^{N \times C_{3D}} \Leftarrow \mathbb{R}^{N \times (C_{2D}+C_{2D})} $), where $[\cdot, \cdot]$ is the concatenation operation. 
With layer normalization~\cite{ba2016layer} ($\text{LN}$), we apply $3\times 3$ depth-wise convolution to encode spatial and channel information with queries $Q_{pri}^{3D}=W_d^Q \text{LN}(X_{pri}^{3D})$, keys $K_{aux}^{3D}=W_d^K \text{LN}(Y_{aux}^{3D})$ and values $V_{aux}^{3D}=W_d^V \text{LN}(Y_{aux}^{3D})$, respectively, obtaining the cross-modal self-attention as:
\begin{equation}
\begin{aligned}
\label{cross_modal_attention}
\mathbf{Attention}(Q, K, V) = & V \mathbf{Softmax} \left( \frac{Q^T K}{\tau} \right), 
\end{aligned}
\end{equation}
where $\tau$ is a learnable scaling factor.
We input $Q_{pri}^{3D}$, $K_{aux}^{3D}$, $V_{aux}^{3D}$ to the above attention module, and the resultant attention map with dimension $\mathbb{R}^{C_{3D} \times C_{3D}}$ is much smaller and more efficient than the original implementation~\cite{dosovitskiy_imagetransformer_iclr_2020} with dimension $\mathbb{R}^{HW \times HW}$ adopted from~\cite{zamir_restormer_CVPR_2022}.
With the cross-modal attention in Eq.~\ref{cross_modal_attention}, we obtain the fused feature ${X_{pri}^{3D}}'$ corresponding to the primary modality $e_{pc}$ by another $1 \times 1$ convolution projection and residual connection as: 
\begin{equation}
\begin{aligned}
\label{equ:threed_fusion}
{X_{pri}^{3D}}' \!= & W_p \mathbf{Attention}(Q_{pri}^{3D}\!, K_{aux}^{3D}, V_{aux}^{3D}) \!+ \!X_{pri}^{3D}.
\end{aligned}
\end{equation}

\noindent\textbf{Multimodal Attention Fusion in 2D branch (MAF RPE-2D).} 
The fusion process in 2D is similar to the 3D branch above. 
To project the sparse point feature into a dense feature at the image plane, we adopt a learnable fusion-aware interpolation~\cite{scene:liu_camliflow_cvpr_2022} and get the projected point feature $e_{pc}^{pj} \in \mathbb{R}^{H \times W \times C_{3D}}$.
Then we regard the image feature as the primary modality feature $X_{pri}^{2D}=e_r$, then the aligned auxiliary feature $\! Y_{aux}^{2D}\! =\! W_a^{2D} \! \left[ e_{pc}^{pj}, e_{ev} \right] $. 
After similar cross-attention, the fused feature ${X_{pri}^{2D}}'$ is obtained as: 
\begin{equation}
\begin{aligned}
\label{equ:twod_fusion}
{X_{pri}^{2D}}' \!= & W_p \mathbf{Attention}(Q_{pri}^{2D}\!, K_{aux}^{2D}, V_{aux}^{2D}) \!+ \!X_{pri}^{2D}. 
\end{aligned}
\end{equation}

\subsection{Mutual Information Regularization (MIR)}
\label{sec:mutual_info_reg}
We incorporate mutual information minimization as a regularizer to explicitly model the cross-modal dependency following disentangled representation learning~\cite{Isolating_Sources} based on the observation that each modality, \ie~RGB image, point cloud and event data, contributes partially to the output, and mutual information minimization is suitable for our task to explore the complementary information of each modality.

We start with the case of two modalities with RGB image and event feature embeddings $e_r$ and $e_{ev}$. Note that these feature embeddings are the features that have been projected into the same spatial space (2D or 3D) in Sec.~\ref{sec:attention}.

To explicitly model the cross-modal correlation, we define cross-modal mutual information between $e_r$ and $e_{ev}$ as $I(e_r; e_{ev}) =  \mathbb{E}_{p(e_r, e_{ev})}\left[\log\frac{p(e_{ev}|e_r)}{p(e_{ev})}\right]$, where $p(e_r, e_{ev})$ is the joint distribution, $p(e_{ev}|e_r)$ is the conditional distribution and $p(e_{ev})$ is the marginal distribution.
With importance sampling~\cite{burda2015importance}, we introduce a variational marginal approximation $q(e_{ev})$ with a variational upper bound $I^{vub}$~\cite{cheng2020club} of mutual information $I(e_r; e_{ev})$ as:
\begin{equation}
\begin{aligned}
I(e_r; \!e_{ev}\!) \!&\leq \!\mathbb{E}_{p(e_r, e_{ev})}\!\left[\log\!\frac{p(e_{ev}|e_r)}{q(e_{ev})}\!\right]\\
&=\!D_{KL}(p(e_{ev}|e_r)\|q(e_{ev})) = I^{vub} = \mathcal{L}_{\text{mi}},
\end{aligned}
\label{equ:mi_loss}
\end{equation}
where $q(e_{ev})$ can be fixed as a standard normal distribution~\cite{alemi2017deep}, \ie~$q(e_{ev})=\mathcal{N}(e_{ev};0,\mathbf{I})$, and $p(e_{ev}|e_r)$ can be modeled with the reparameterization trick~\cite{VAE1}, thus the Kullback-Leibler (KL) divergence term $D_{KL}$ within $I^{vub}$ can be solved in closed form.

In the case of three modalities feature embeddings $e_r$, $e_{pc}$ and $e_{ev}$, respectively, the interaction information $II(e_r; e_{pc}; e_{ev})$~\cite{interaction_information}, as a multivariate generalization of the mutual information, is upper bounded by:
\begin{equation}
\begin{aligned}
&\!II(e_r;\!e_{pc};\!e_{ev}) \! \leq \! \min \{I(e_r;\!e_{pc}),\!I(e_{pc};\!e_{ev}),\!I(e_r;\!e_{ev})\}\!\\
&\leq \min \{I^{vub}(e_r;e_{pc}),I^{vub}(e_{pc};e_{ev}),I^{vub}(e_r;e_{ev})\}.
\end{aligned}
\end{equation}

To compute $I^{vub}$, we need to design a transition function, achieving the transformation of one modality to the other and assume the variational marginal approximation $q$ as the standard normal distribution for closed-form KL divergence computation. In practice, the standard normal distribution assumption of $q$ leads to high-bias mutual information estimation. Alternatively, we first map the representation of each modality ($e_r, e_{pc}, e_{ev}$) to a common manifold, with reparameterization trick~\cite{VAE1} in the end to achieve Gaussian latent code of each modality. Then, we compute KL divergence between the two Gaussian latent codes, leading to the final mutual information regularization term:
\begin{equation}
    \begin{aligned}
\label{equ:ii_loss}
\mathcal{L}_\text{ii} = I^{vub}(e_r;e_{pc})+I^{vub}(e_{pc};e_{ev})+I^{vub}(e_r;e_{ev}),
\end{aligned}
\end{equation}
where we compute the sum of $I^{vub}$ instead of choosing the minimum for stable training.

\subsection{Pyramid Fusion and Joint Estimation Model}
\label{sec:joint_estimation}
With the proposed multimodal attention fusion and the mutual information regularization term, we achieve once multimodal feature fusion. 
Inspired by CamLiFlow~\cite{scene:liu_camliflow_cvpr_2022}, we further perform multi-stage feature fusion implicitly and explicitly.
Here we present the details of multi-stage feature fusion, and more details about the network structure are given in the supplementary materials. 
As shown in Fig.~\ref{fig:network}, our model contains both 2D and 3D branches, where each branch consists of feature extraction, correlation construction and flow estimation. 

In feature extraction, we first voxelized the raw events $E\!=\!\{ x_i, y_i, t_i, p_i \}^K$ into one event voxel $EV\! \in\! \mathbb{R}^{H \times W \times B}$ that can be used as input to our network, where $K$ is the number of events during the period between two frames and $B$ is the manually set number of time intervals to sample events. 
We apply three Siamese encoders to construct feature pyramids (${\{e_{r_1},e_{r_2}}\}_l$, $\{{e_{{pc}_1},e_{{pc}_2}\}}_l$ and $e_{{ev}_l}$ with pyramid layers $l\!\in\! [1, L]$) for three modalities respectively.
As spans between two frames, the event data is not included in \textit{Feature Stage Fusion}, which is applied to fuse the features of two frames RGB images and corresponding point clouds with two-modal attention fusion, \ie simplify the auxiliary feature to a single modality in Eq.~\ref{equ:threed_fusion}, \ref{equ:twod_fusion} and mutual information regularization in Eq.~\ref{equ:mi_loss} for both 2D and 3D branches.

In correlation construction, we first warp the second frame of image and point cloud features using the coarse optical flow and scene flow (initialize with zero) from the previous pyramid layer, then construct correlation motion features by computing 2D and 3D cost volumes
and fused them with the event feature at \textit{Motion Stage Fusion}.
In flow estimation, we construct a motion decoder and flow estimator, and perform an \textit{Estimation Stage Fusion} between them to fuse the hidden motion features from the two branches decoder with the event feature. 
In these two fusion stages, we conduct multimodal attention fusion in Eq.~\ref{equ:threed_fusion}, \ref{equ:twod_fusion} and mutual information regularization in Eq.~\ref{equ:ii_loss} for 2D and 3D motion features and event feature, because events can provide complementary information to enhance the motion correlation construction. 
The estimated optical flow and scene flow are fed into the next pyramid layer to achieve coarse-to-fine predictions. 
We take the optical flow and scene flow from the last pyramid layer as our final joint estimations.

\subsection{Objective Functions}
The objective functions in the training of our model are divided into feature representation loss and task loss.
The former consists of multiple mutual information regularizations at each fusion stage. ${\mathcal{L}_\text{mi}^{fs1}}_l$, ${\mathcal{L}_\text{mi}^{fs2}}_l$ represent the mutual information minimization loss imposed on the RGB image and point cloud features of the first and the second frame at the \textbf{F}eature \textbf{S}tage fusion, and ${\mathcal{L}_\text{ii}^{ms}}_l$ and ${\mathcal{L}_\text{ii}^{es}}_l$ represent on both RGB image, point cloud and event features at the \textbf{M}otion \textbf{S}tage and \textbf{E}stimation \textbf{S}tage fusion. Thus the feature representation loss is the sum of all stages and is weighted at each pyramid level:
\begin{equation}
\begin{aligned}
\label{equ:feature_loss}
\mathcal{L}_\text{feat} = \sum^L_{l=1} \lambda_l \Big[ {\mathcal{L}_\text{mi}^{fs1}}_l + {\mathcal{L}_\text{mi}^{fs2}}_l+ {\mathcal{L}_\text{ii}^{ms}}_l + {\mathcal{L}_\text{ii}^{es}}_l \Big],
\end{aligned}
\end{equation}
where $l$ is the pyramid level, and $\lambda_l$ is used to reweight the contribution of each pyramid level.

The latter task loss measures the $L_2$ distance between the ground-truth and the model output at each pyramid level by: 
\begin{equation}
\begin{aligned}
\label{equ:task_loss}
\mathcal{L}_\text{task} \!= \sum^L_{l=1} \lambda_l \Big[ & \sum_{\mathbf{x}} (\Vert f_l^{pred}(\mathbf{x}) - f_l^{gt}(\mathbf{x})\Vert_2) + \\ 
\alpha & \sum_{\mathbf{p}}(\Vert s_l^{pred}(\mathbf{p}) - s_l^{gt}(\mathbf{p})\Vert_2) \Big] , 
\end{aligned}
\end{equation}
where $\mathbf{x}$, $\mathbf{p}$ are the valid image positions and point coordinates, $f_l^{pred}$, $s_l^{pred}$ are the estimated 2D optical flow and 3D scene flow and $f_l^{gt}$, $s_l^{gt}$ are the corresponding resized ground truth at the $l$-th pyramid level, respectively.
$\alpha$ is the weight to balance 2D and 3D errors.

The total loss is a weighted sum of the above: 
\begin{equation}
\begin{aligned}
\label{equ:total_loss}
\mathcal{L} = \mathcal{L}_\text{task} + \beta \mathcal{L}_\text{feat}, 
\end{aligned}
\end{equation}
where $\beta$ is the weight to balance two losses in training.

\begin{figure*}[tbp]
\small
\centering
\begin{minipage}[t]{0.163\linewidth}
\centering
\includegraphics[width=\textwidth]{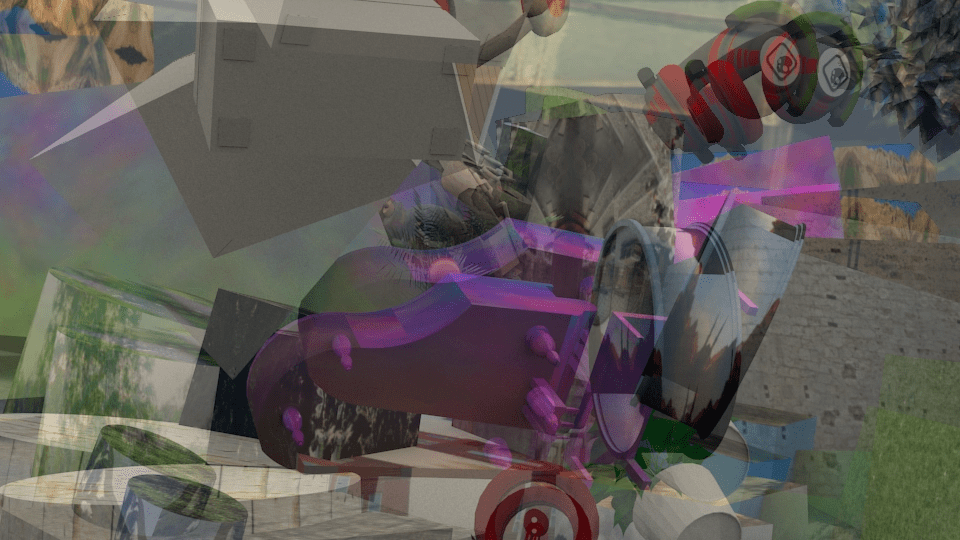} \\
\vspace{0.01cm} 
\includegraphics[width=\textwidth]{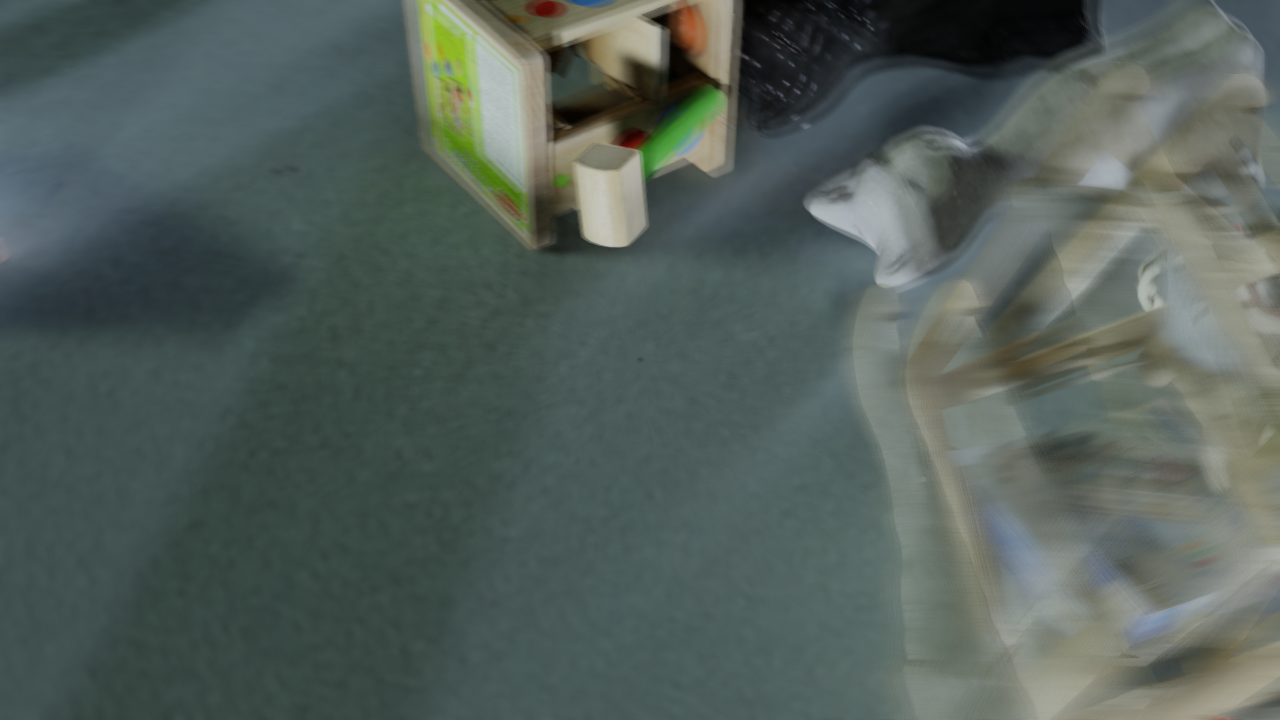} \\
\vspace{0.01cm} 
\includegraphics[width=\textwidth]{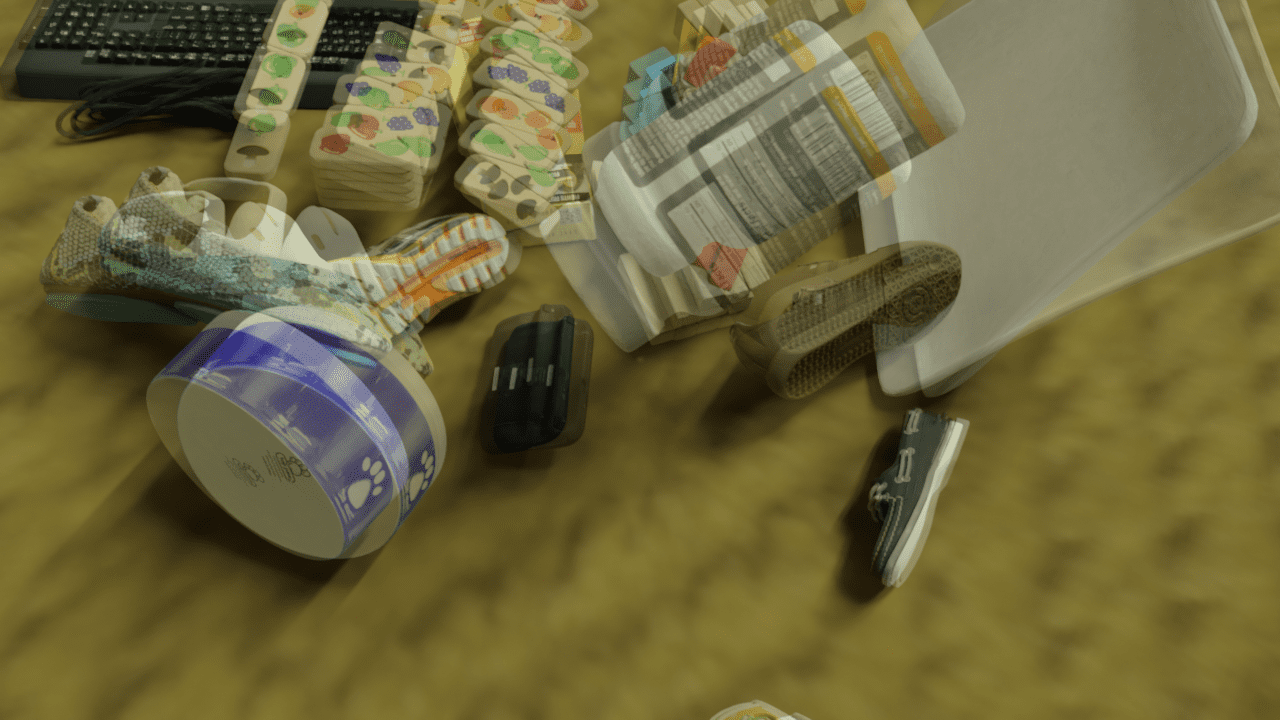} \\
\vspace{-0.1cm}
Overlapped Frames
\end{minipage}%
\hspace{0.002cm}
\begin{minipage}[t]{0.163\linewidth}
\centering
\includegraphics[width=0.98\textwidth, frame]{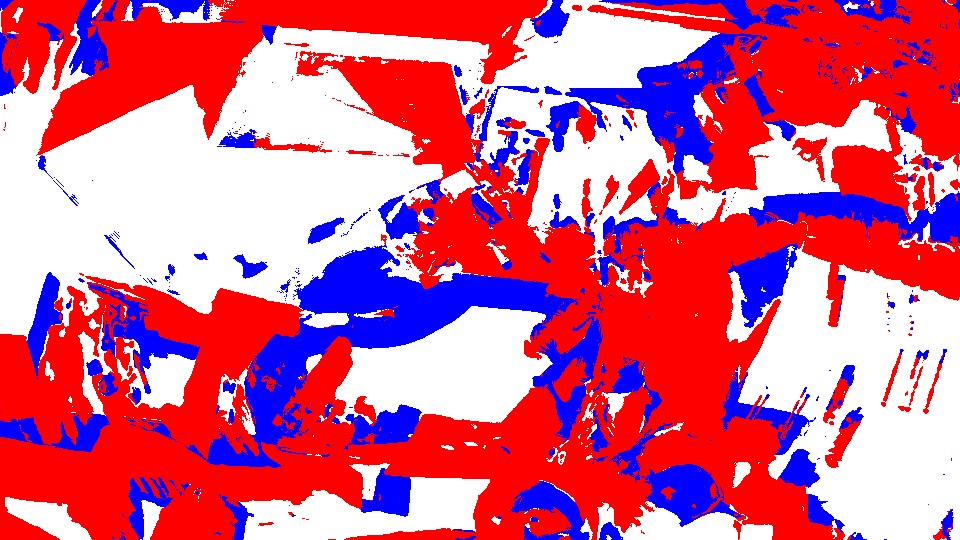} \\
\vspace{0.01cm} 
\includegraphics[width=0.98\textwidth, frame]{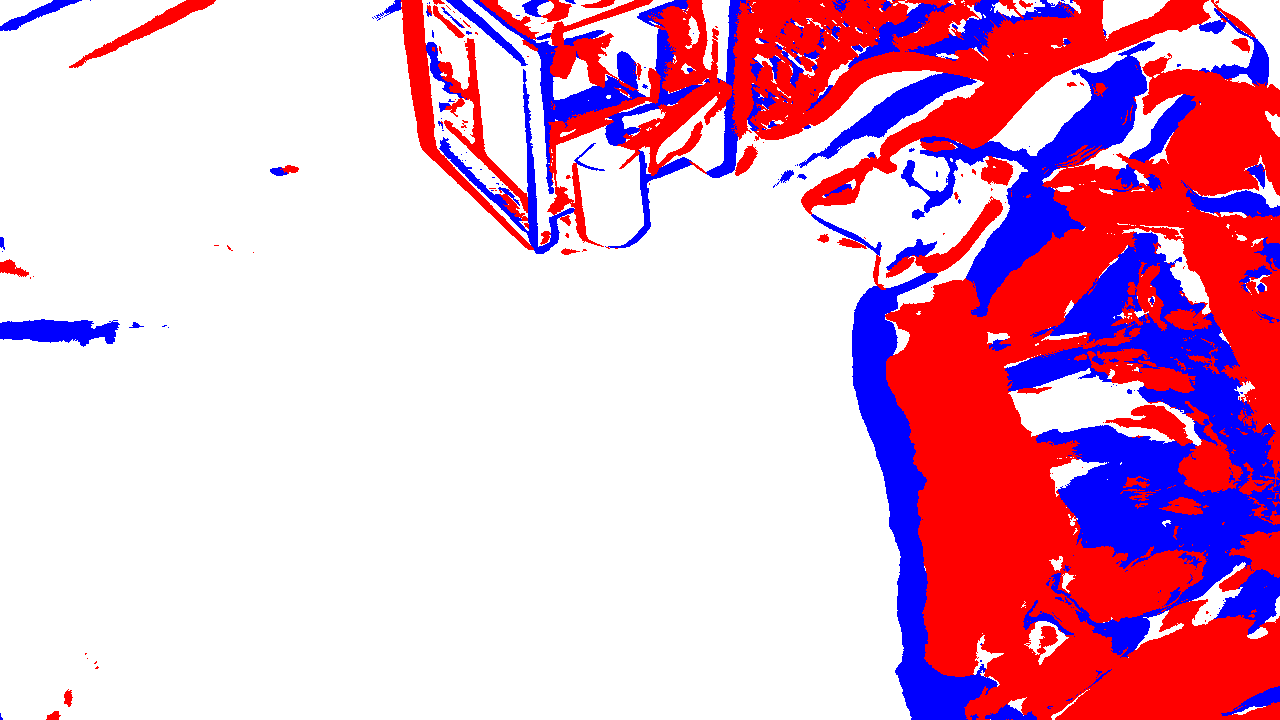} \\
\vspace{0.01cm} 
\includegraphics[width=0.98\textwidth, frame]{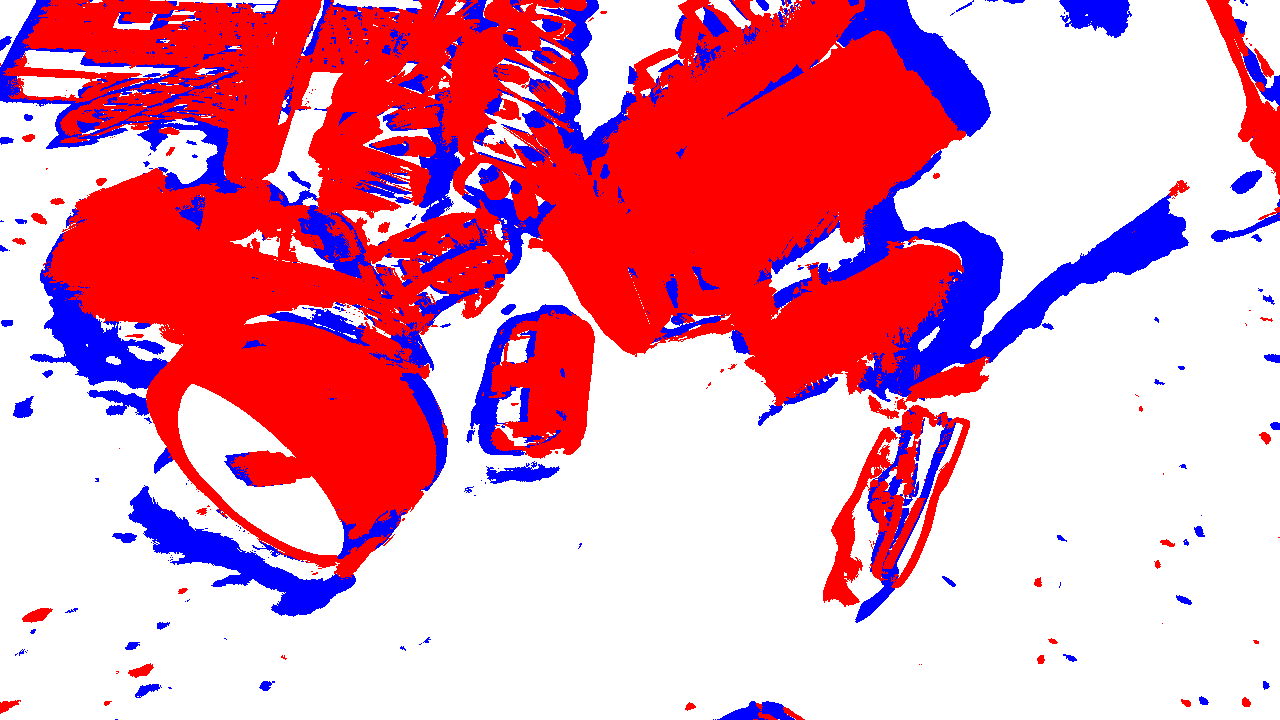} \\
\vspace{-0.1cm}
Events
\end{minipage}%
\hspace{0.002cm}
\begin{minipage}[t]{0.163\linewidth}
\centering
\includegraphics[width=\textwidth]{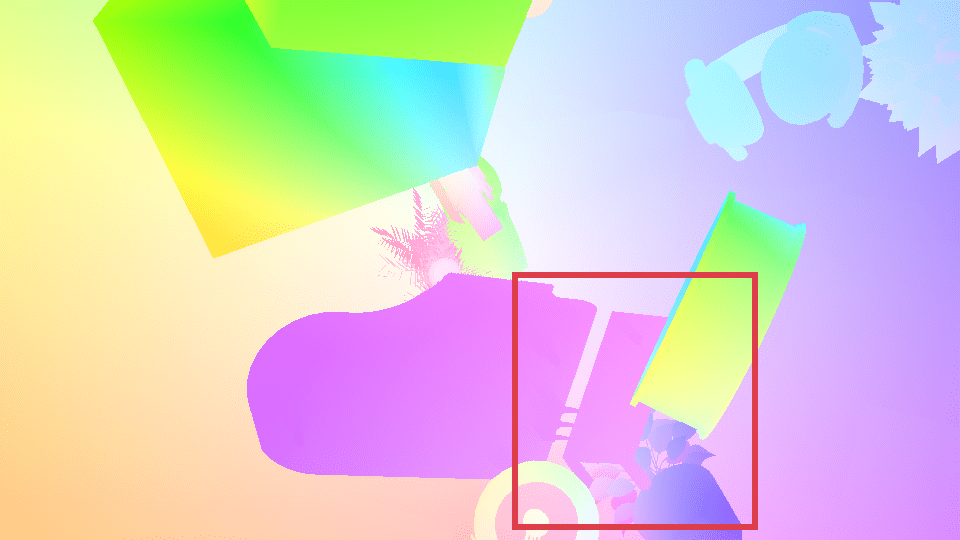} \\
\vspace{0.01cm} 
\includegraphics[width=\textwidth]{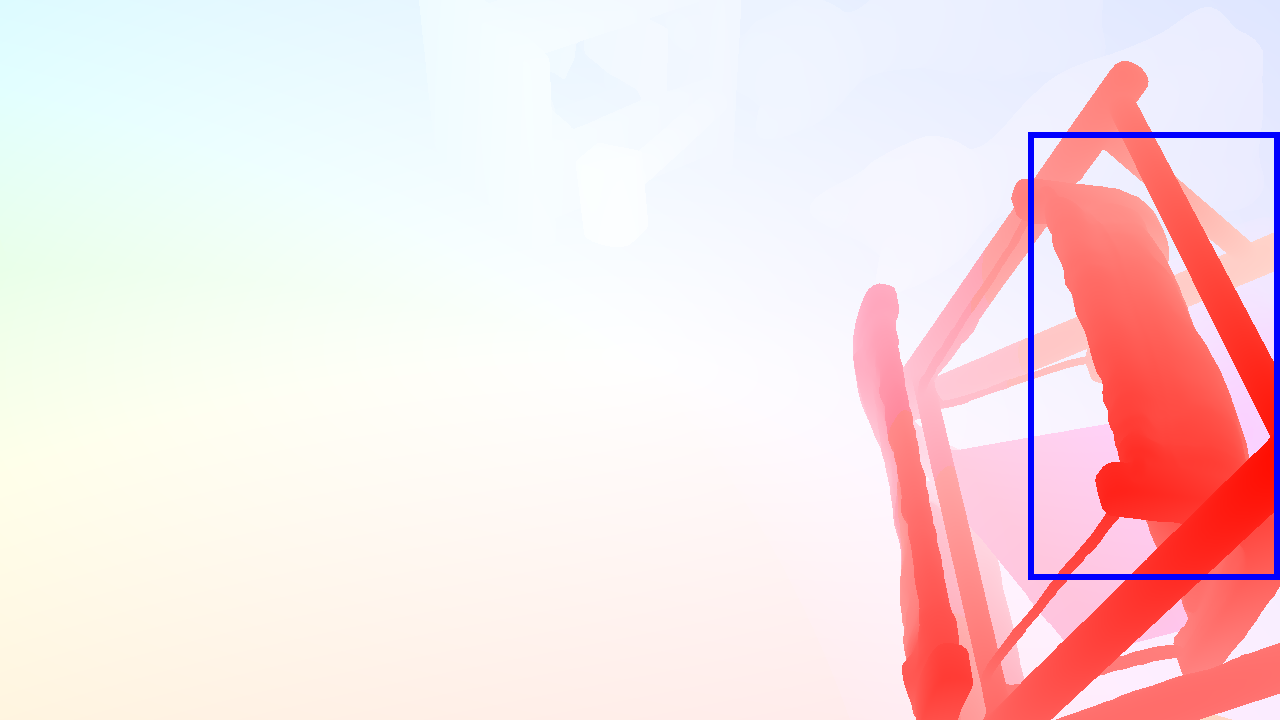} \\
\vspace{0.01cm} 
\includegraphics[width=\textwidth]{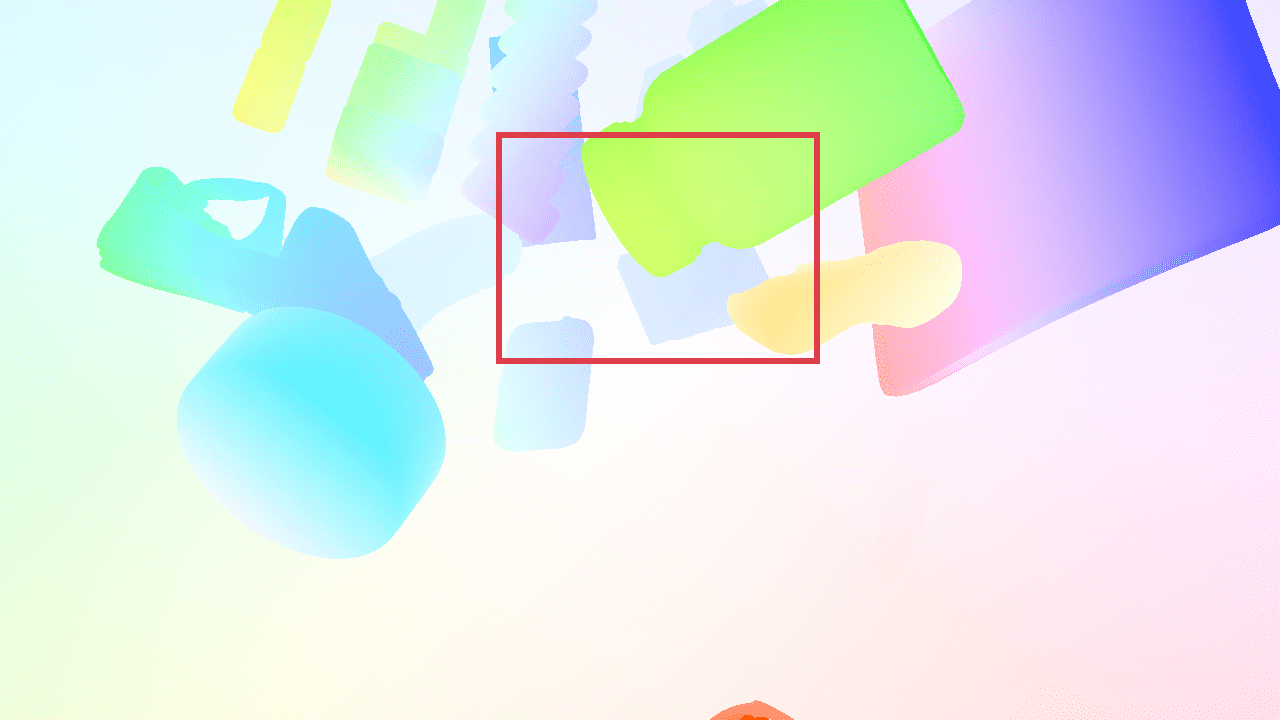} \\
\vspace{-0.1cm}
Optical Flow GT
\end{minipage}%
\hspace{0.002cm}
\begin{minipage}[t]{0.163\linewidth}
\centering
\includegraphics[width=\textwidth]{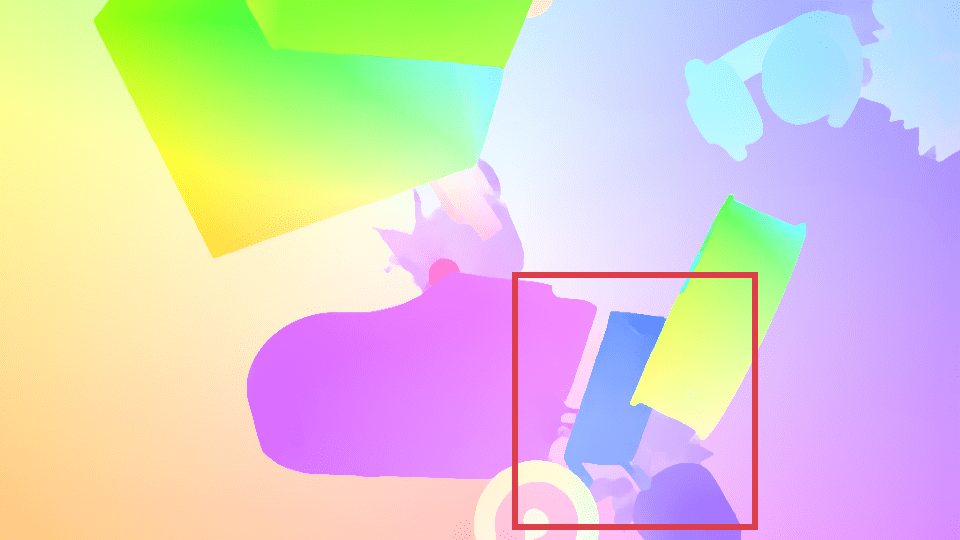} \\
\vspace{0.01cm} 
\includegraphics[width=\textwidth]{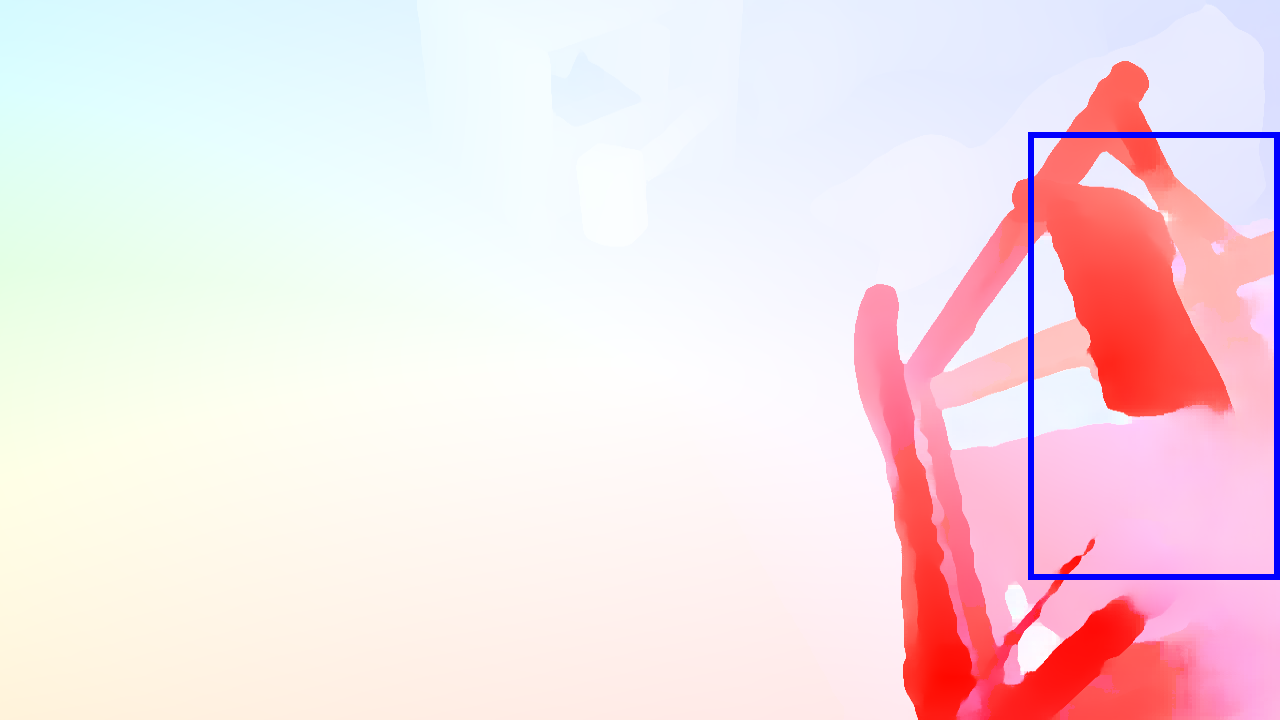} \\
\vspace{0.01cm} 
\includegraphics[width=\textwidth]{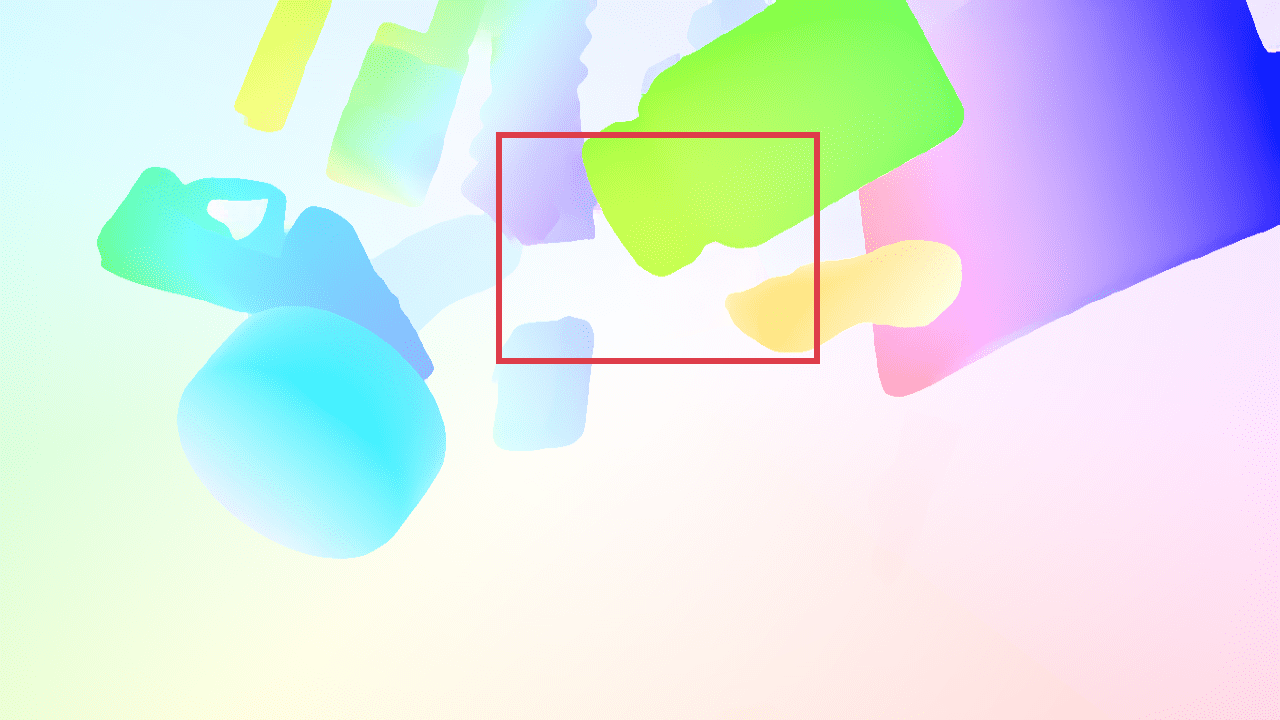} \\
\vspace{-0.1cm}
CamLiFlow~\cite{scene:liu_camliflow_cvpr_2022}
\end{minipage}%
\hspace{0.002cm}
\begin{minipage}[t]{0.163\linewidth}
\centering
\includegraphics[width=\textwidth]{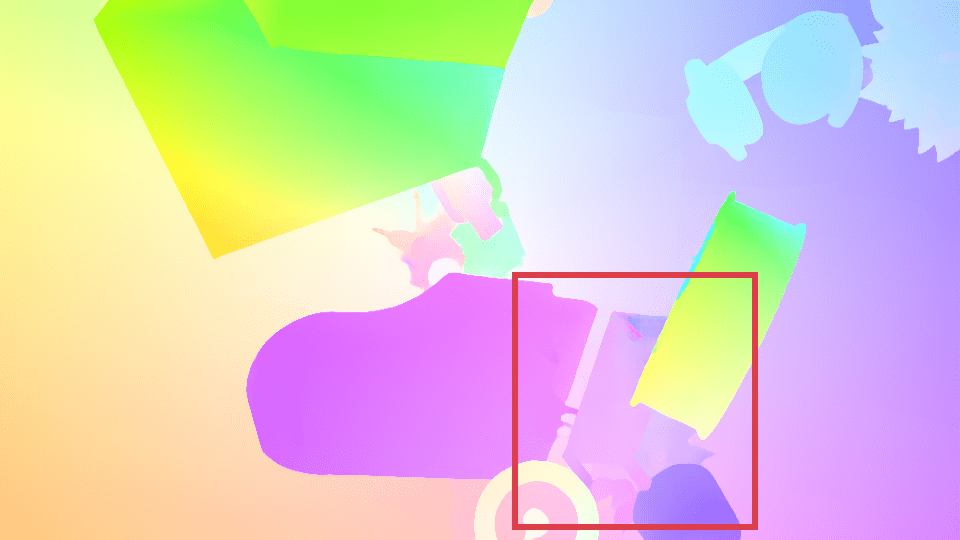} \\
\vspace{0.01cm} 
\includegraphics[width=\textwidth]{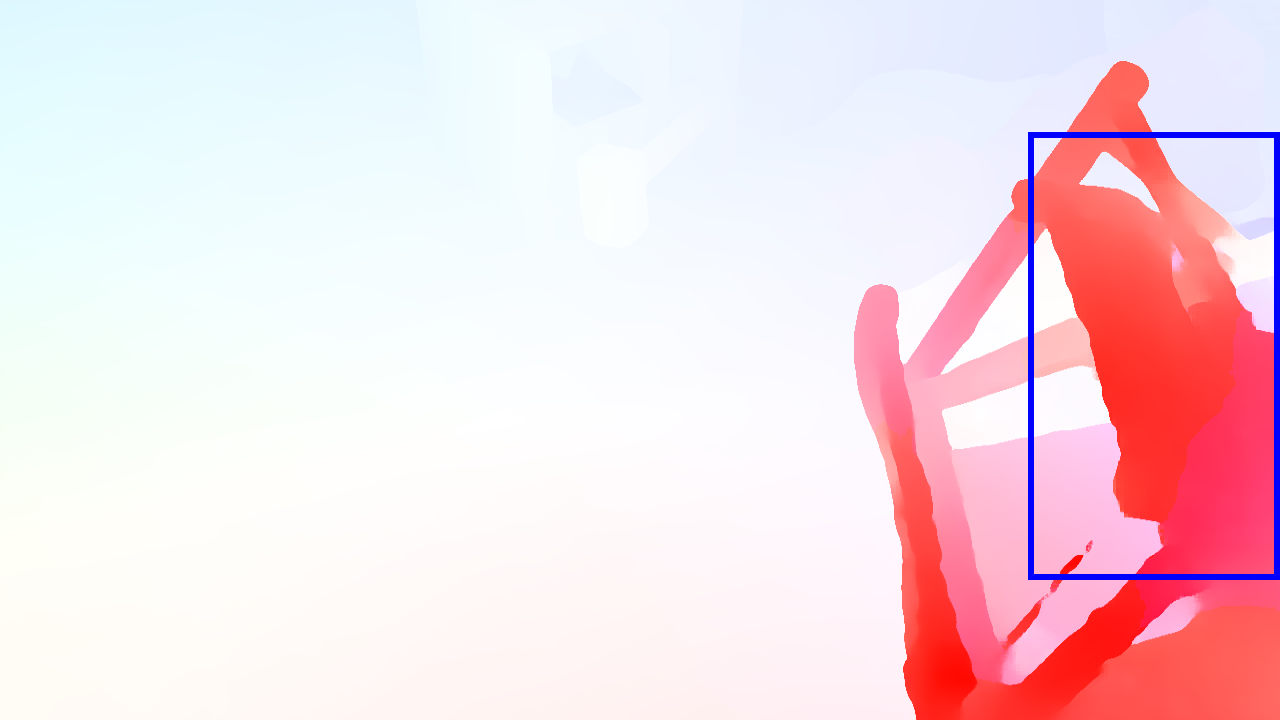} \\
\vspace{0.01cm} 
\includegraphics[width=\textwidth]{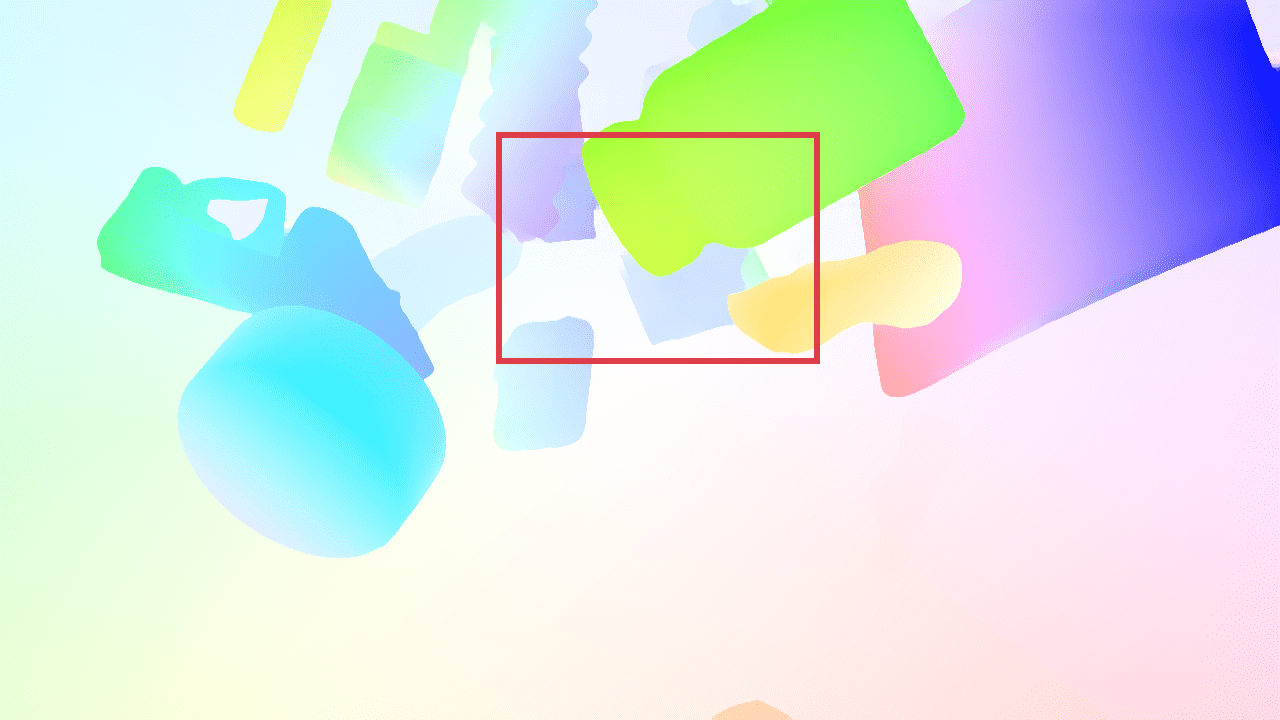} \\
\vspace{-0.1cm}
CamLiFlow+Events
\end{minipage}%
\hspace{0.002cm}
\begin{minipage}[t]{0.163\linewidth}
\centering
\includegraphics[width=\textwidth]{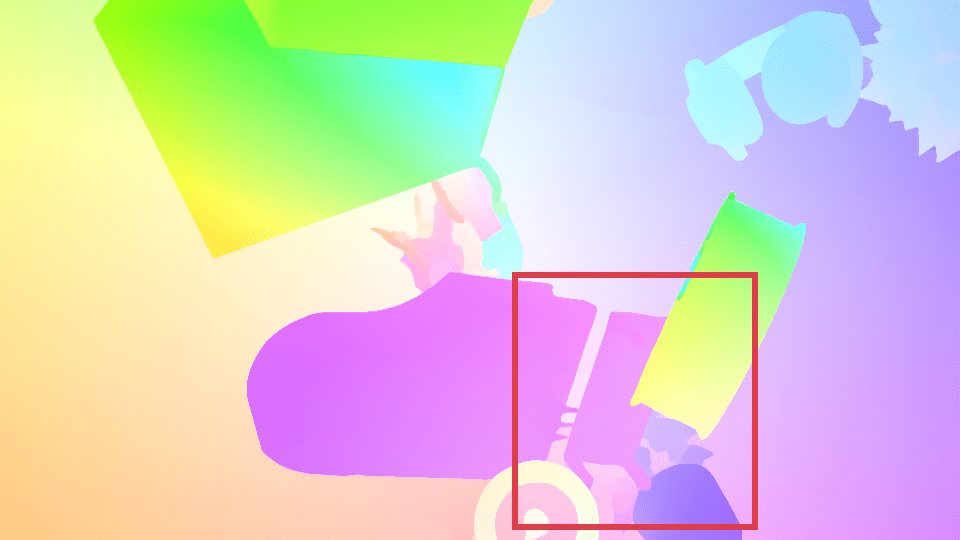} \\
\vspace{0.01cm} 
\includegraphics[width=\textwidth]{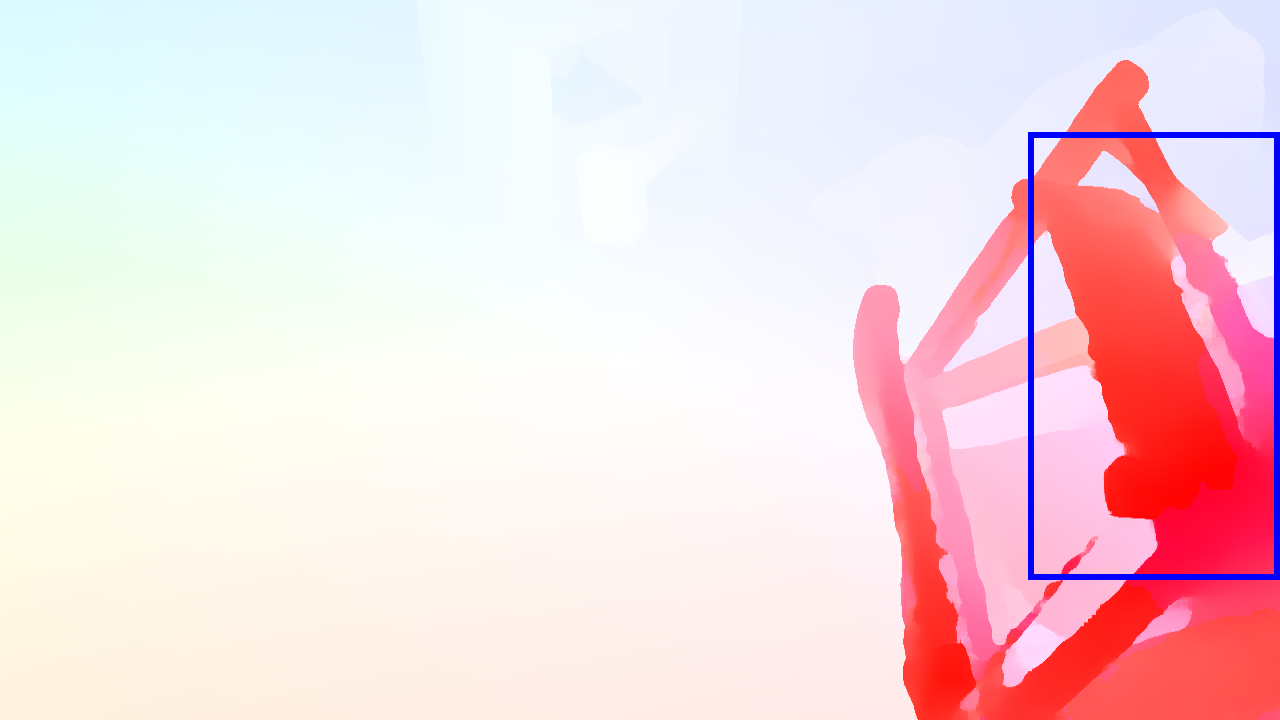} \\
\vspace{0.01cm} 
\includegraphics[width=\textwidth]{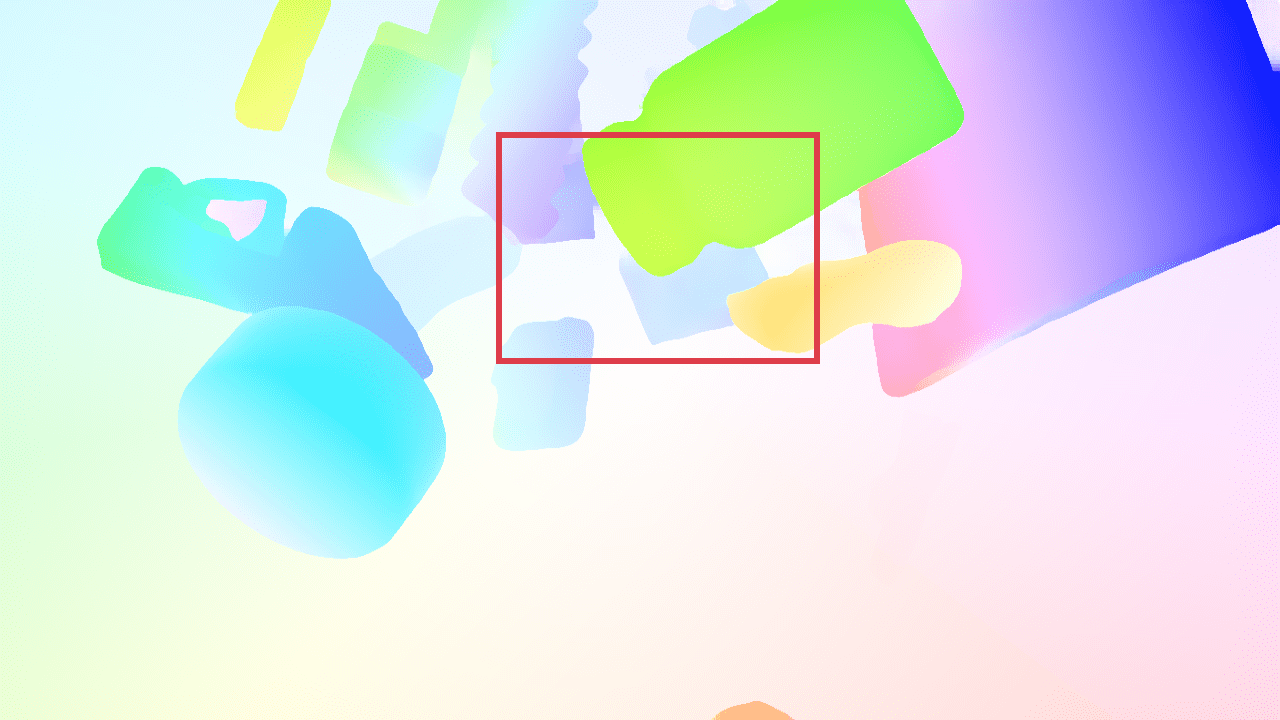} \\
\vspace{-0.1cm}
RPEFlow~(Ours)
\end{minipage}%

\begin{minipage}[t]{0.185\linewidth}
\centering
\includegraphics[width=0.96\textwidth, frame]{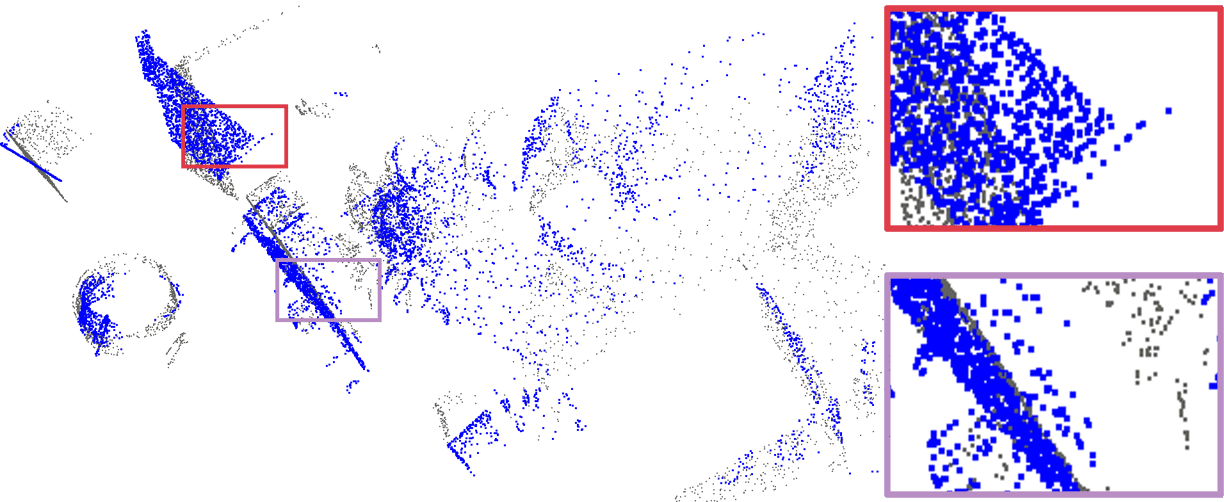} \\
\vspace{0.01cm} 
\includegraphics[width=0.96\textwidth, frame]{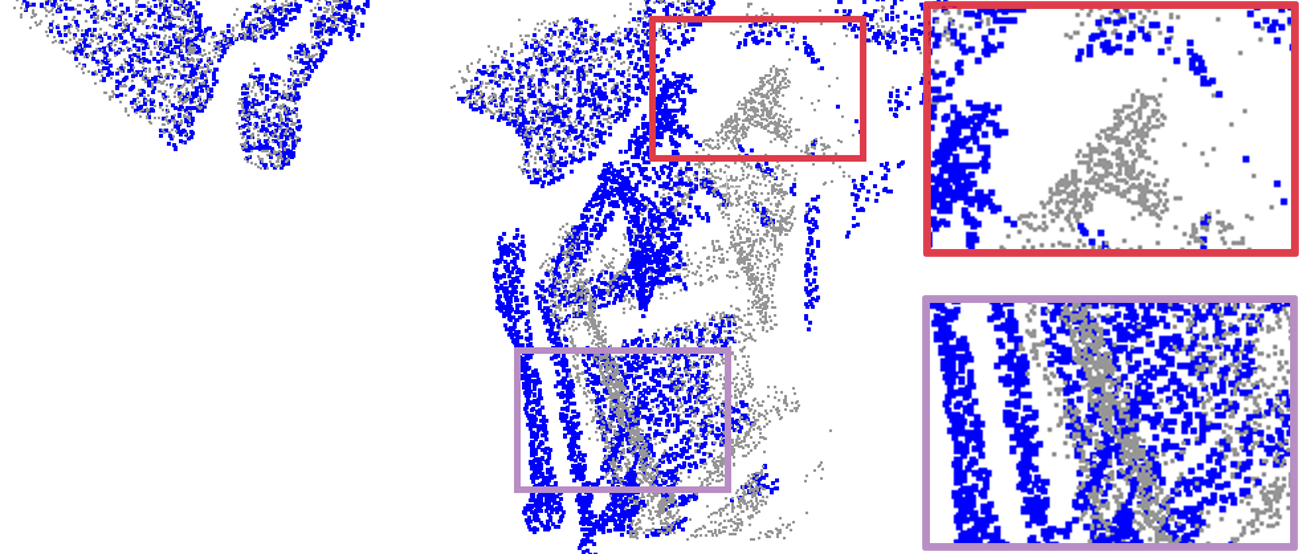} \\
\vspace{0.01cm} 
\includegraphics[width=0.96\textwidth, frame]{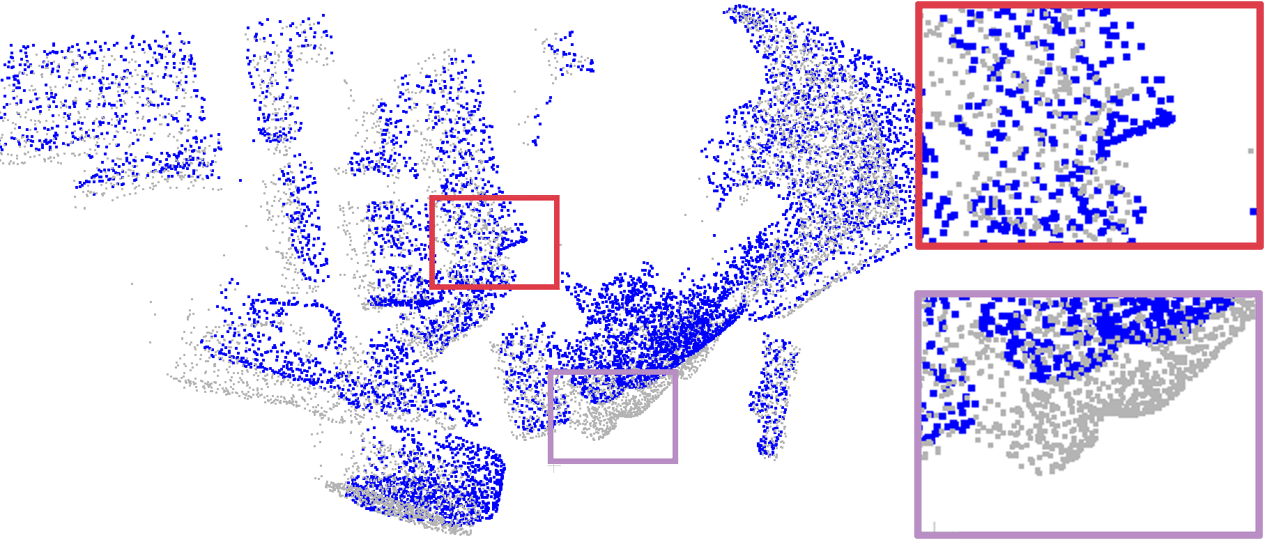} \\
\vspace{-0.1cm}
Point Clouds
\end{minipage}%
\hspace{0.005\linewidth}
\begin{minipage}[t]{0.185\linewidth}
\centering
\includegraphics[width=0.96\textwidth, frame]{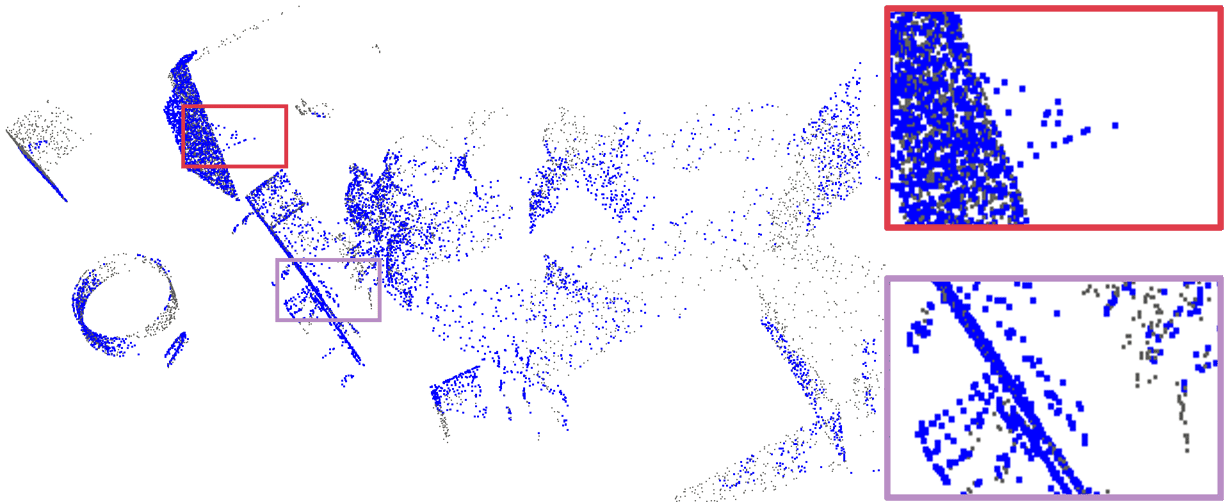} \\
\vspace{0.01cm} 
\includegraphics[width=0.96\textwidth, frame]{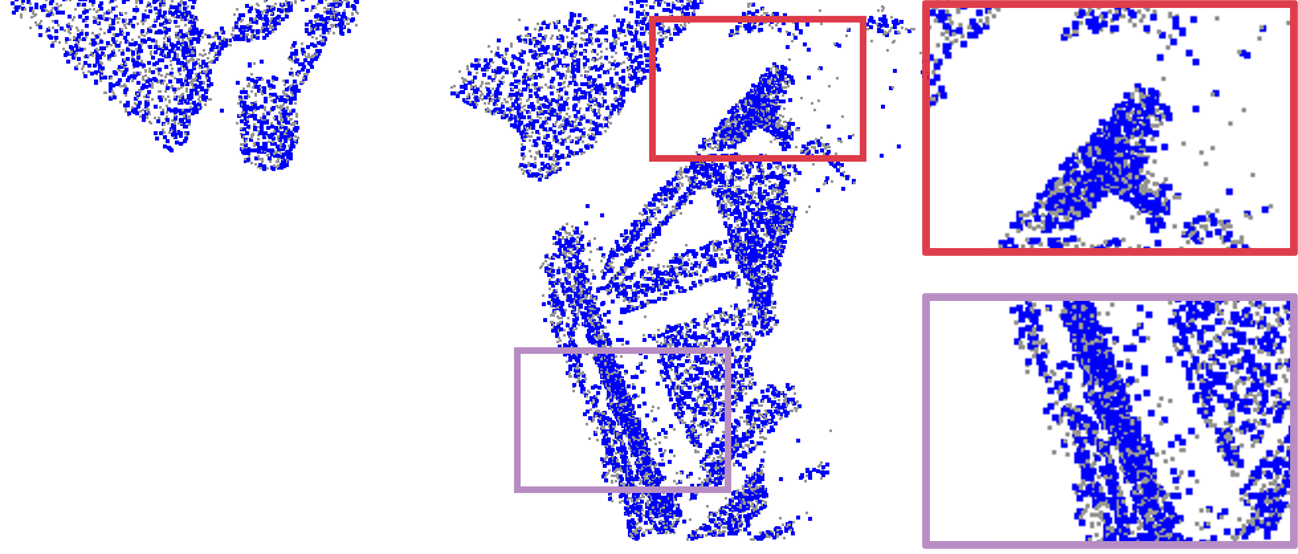} \\
\vspace{0.01cm} 
\includegraphics[width=0.96\textwidth, frame]{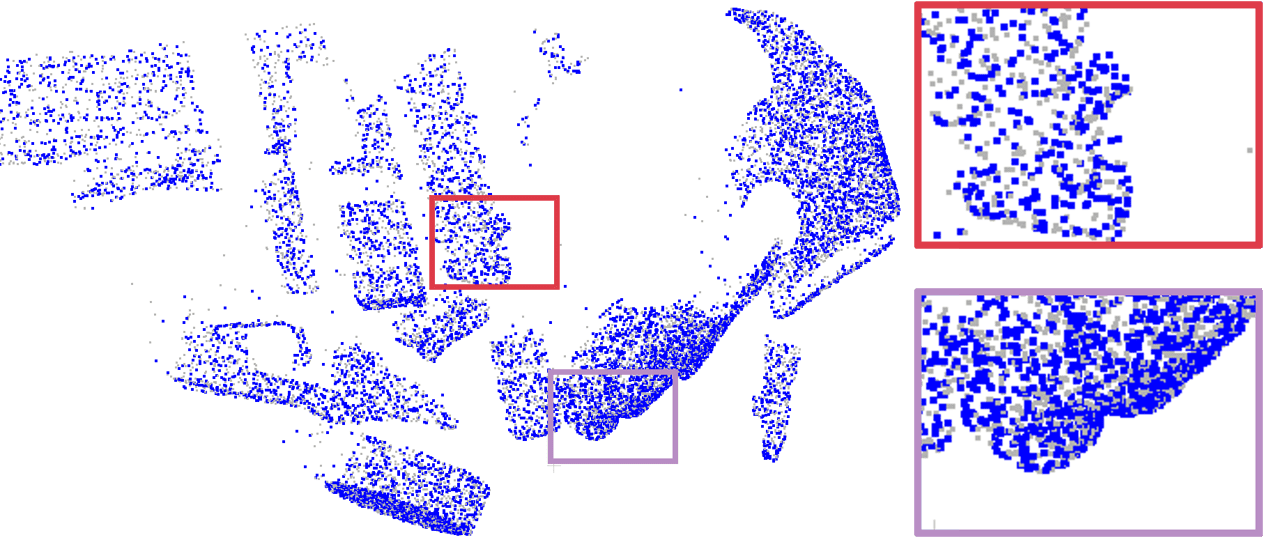} \\
\vspace{-0.1cm} 
Scene Flow GT
\end{minipage}%
\hspace{0.005\linewidth}
\begin{minipage}[t]{0.185\linewidth}
\centering
\includegraphics[width=0.96\textwidth, frame]{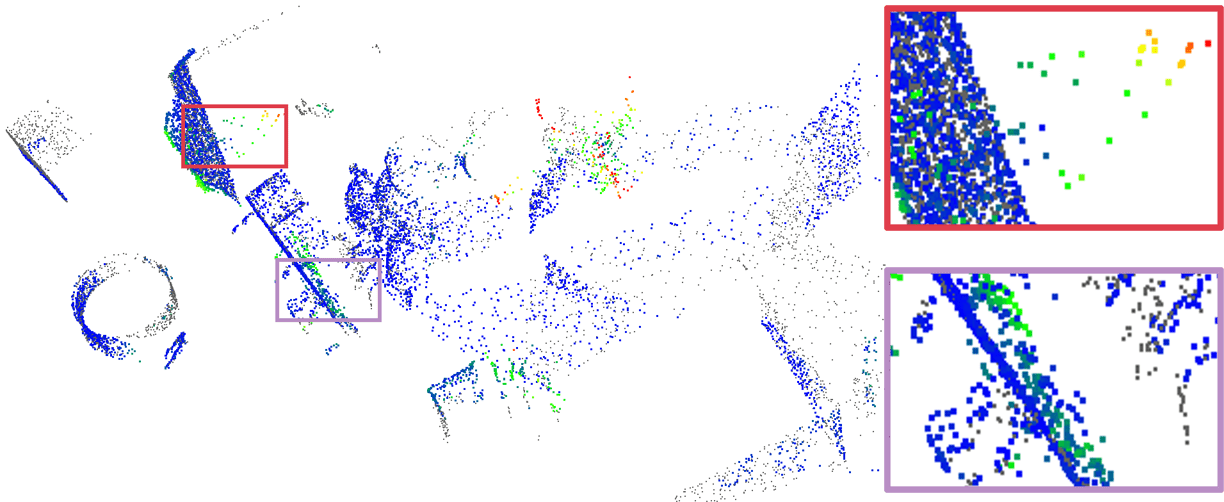} \\
\includegraphics[width=0.96\textwidth, frame]{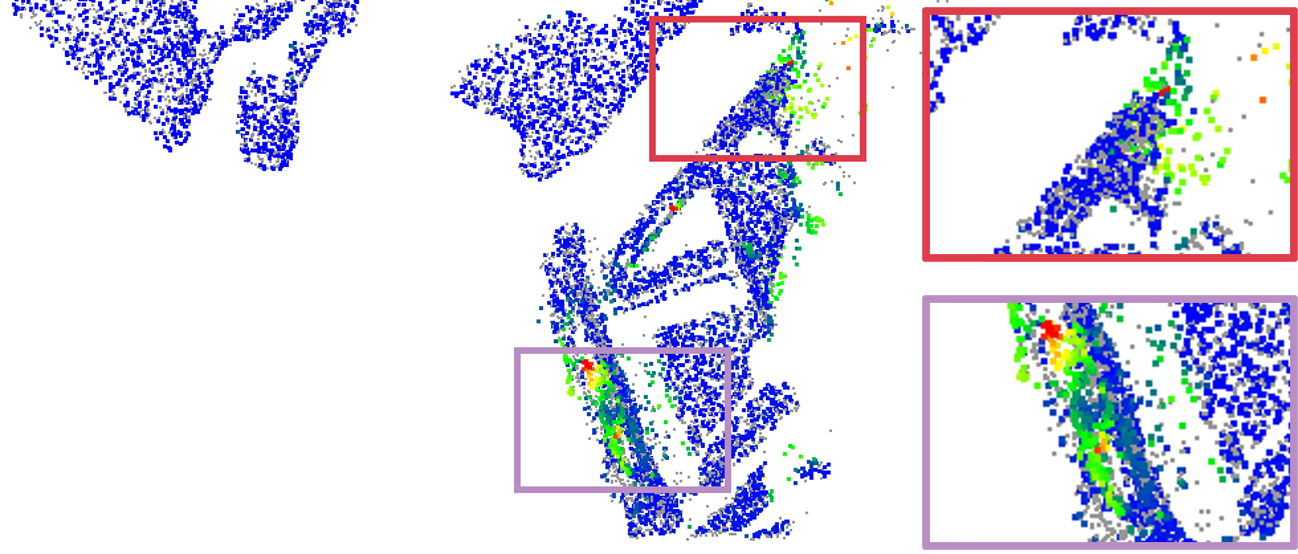} \\
\vspace{0.01cm} 
\includegraphics[width=0.96\textwidth, frame]{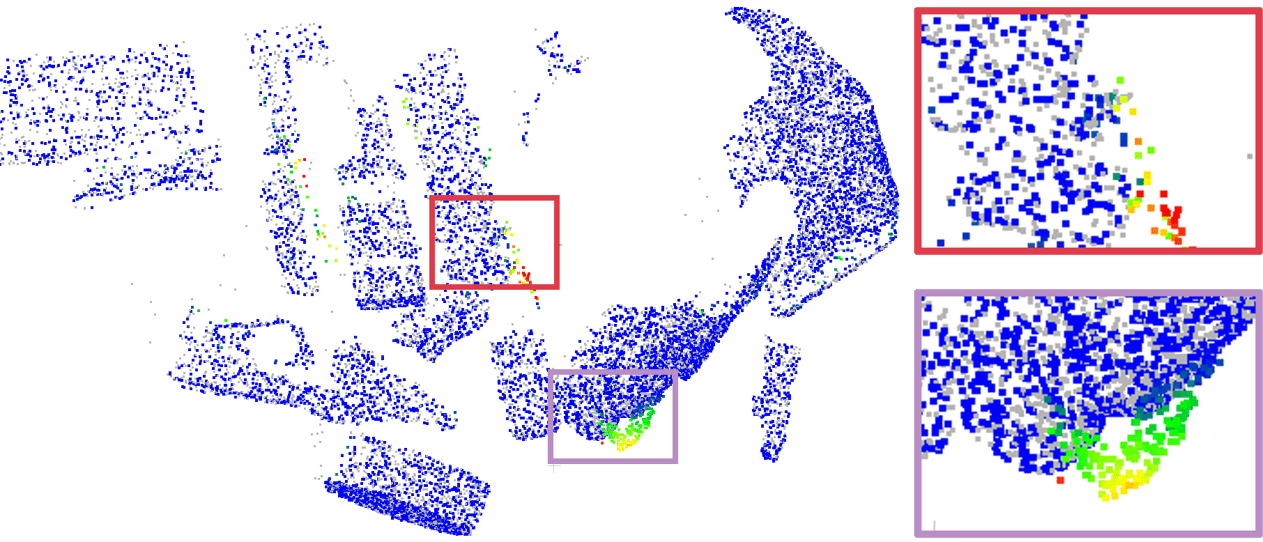} \\
\vspace{-0.1cm}
CamLiFlow~\cite{scene:liu_camliflow_cvpr_2022}
\end{minipage}%
\hspace{0.005\linewidth}
\begin{minipage}[t]{0.185\linewidth}
\centering
\includegraphics[width=0.96\textwidth, frame]{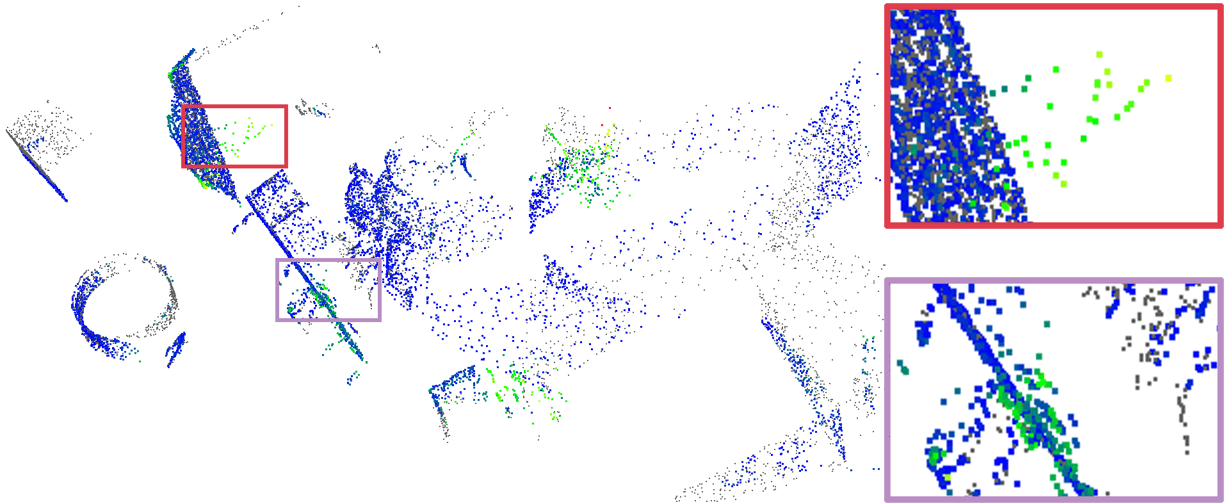} \\
\includegraphics[width=0.96\textwidth, frame]{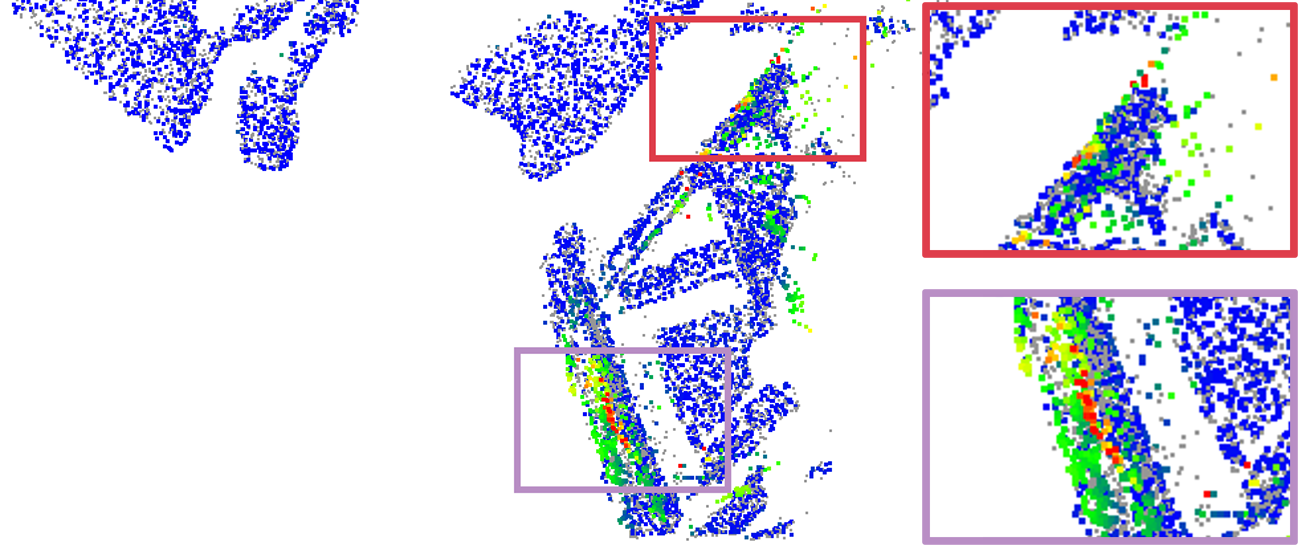} \\
\vspace{0.01cm} 
\includegraphics[width=0.96\textwidth, frame]{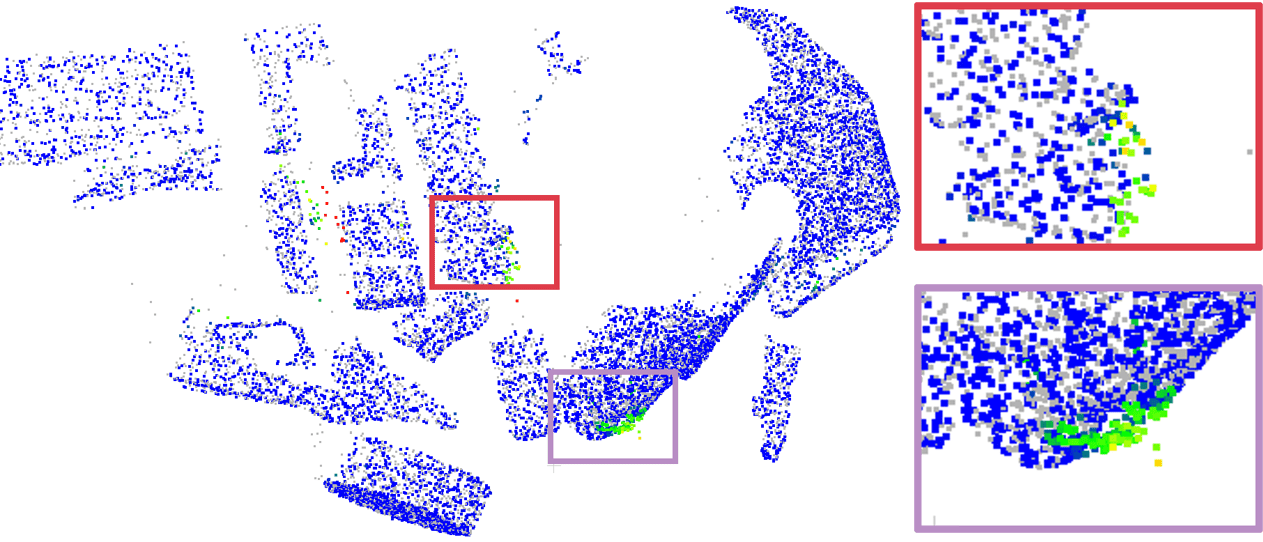} \\
\vspace{-0.1cm}
CamLiFlow+Events
\end{minipage}%
\hspace{0.005\linewidth}
\begin{minipage}[t]{0.185\linewidth}
\centering
\includegraphics[width=0.96\textwidth, frame]{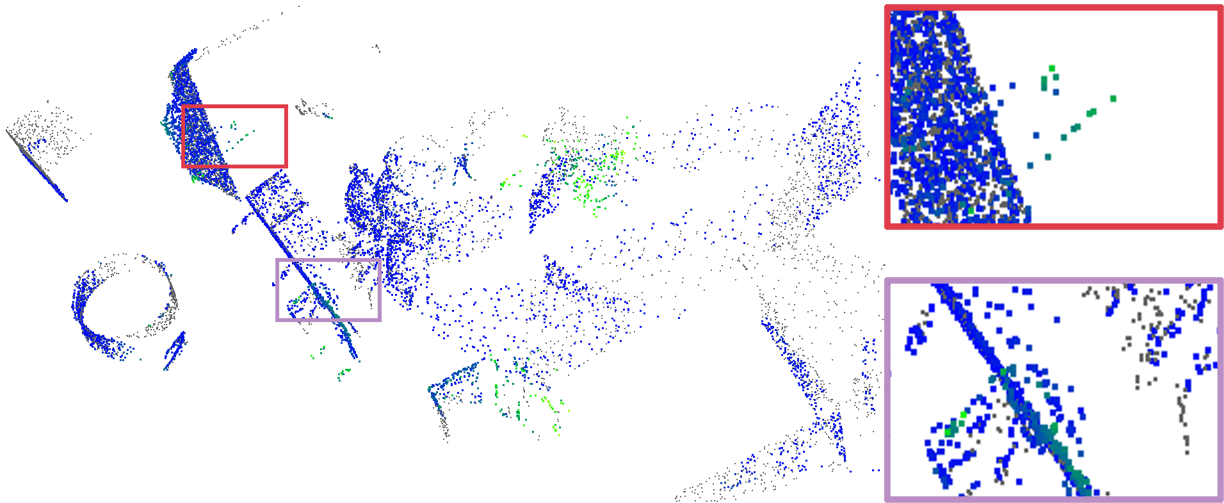} \\
\includegraphics[width=0.96\textwidth, frame]{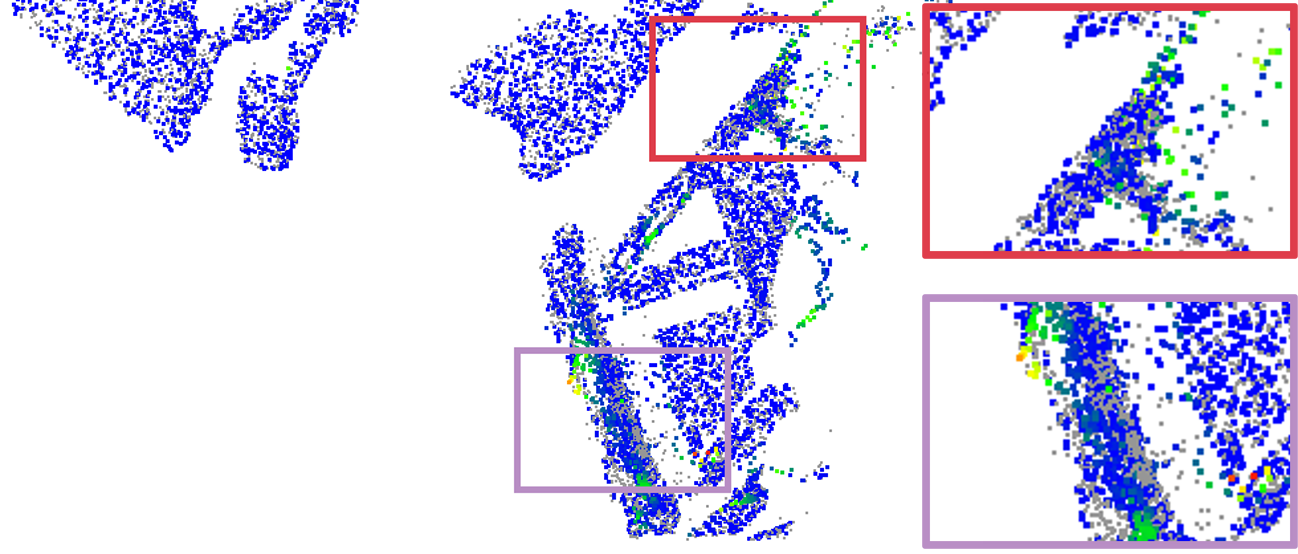} \\
\vspace{0.01cm} 
\includegraphics[width=0.96\textwidth, frame]{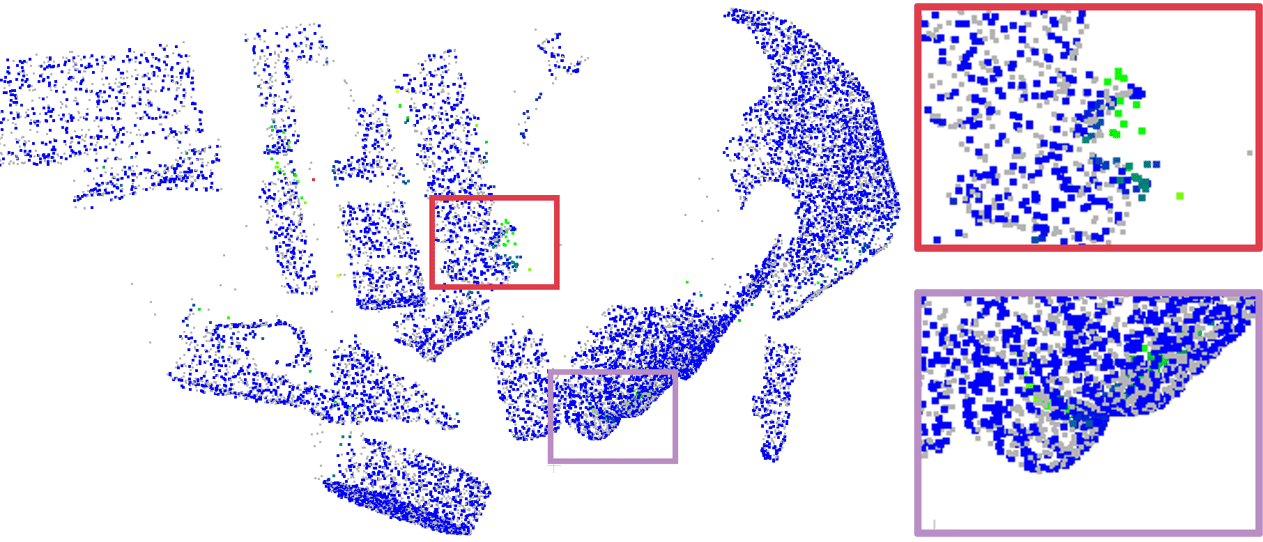} \\
\vspace{-0.1cm}
RPEFlow~(Ours)
\end{minipage}%
\begin{minipage}[t]{0.001\linewidth}
\centering
\raisebox{-0.7\height}{\includegraphics[height=3.5cm]{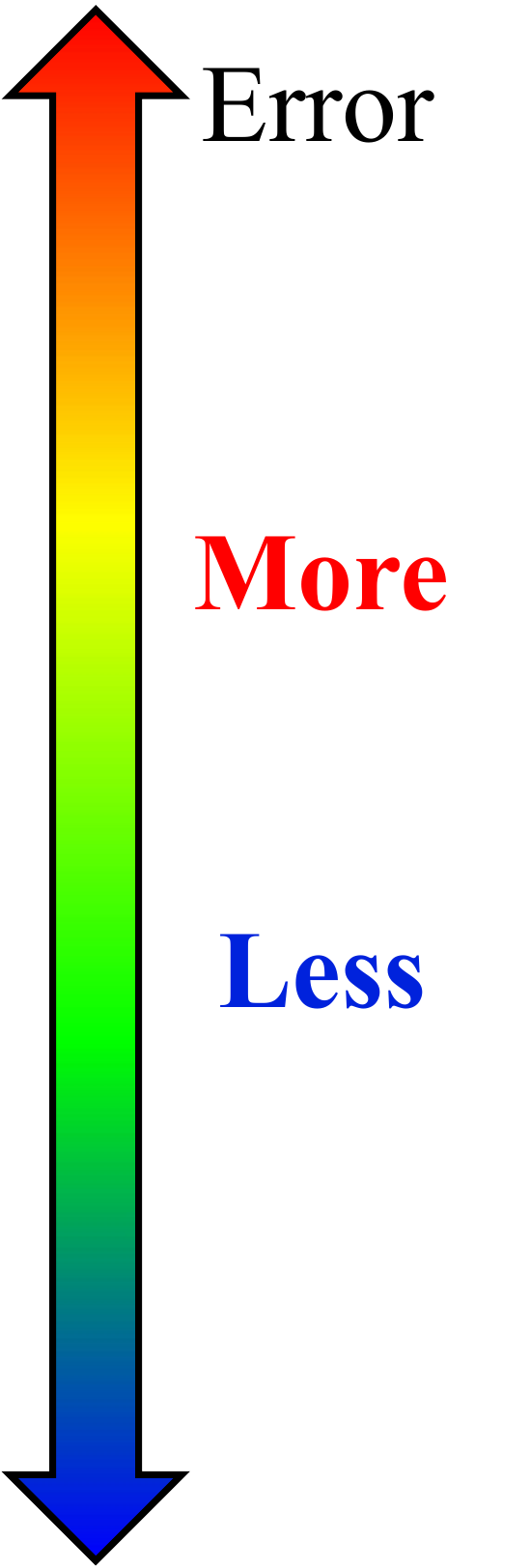}}
\end{minipage}%

\centering
\caption{\textbf{Visual comparisons} on simulated data, in which the top row is on the ``val'' split of FlyingThings3D~\cite{flowdatasets:Mayer_FlyingThing3d_CVPR_2016} dataset and the reset two are on the test split of our simulated EKubric dataset.
For the bottom 3D comparisons, blue indicates a lower error, red indicates a higher error, and green indicates the median. 
Best to zoom in on the screen for detailed comparisons.
} 
\label{viz:things_kubric}
\end{figure*}

\begin{table*}[tbp]
    \caption{\textbf{Performance comparison} on the ``val'' split of the FlyingThings3D~\cite{flowdatasets:Mayer_FlyingThing3d_CVPR_2016} subset.}
    \renewcommand\arraystretch{0.75}
    \centering
    \setlength{\tabcolsep}{2.8mm}{
    \small
    \begin{tabular}{cccccccc}
    \toprule
    Input & Method & $\rm {EPE}_{2D}$ & $\rm {ACC}_{1px}$ & $\rm {EPE}_{3D}^{N.Occ}$ & $\rm {ACC}_{.05}^{N.Occ}$ & $\rm {EPE}_{3D}^{Full}$ & $\rm {ACC}_{.05}^{Full}$ \\
    \midrule
    \multirow{2}*{RGB} 
    & RAFT~\cite{flow:Teed_RAFT_ECCV_2020} & 3.12 & 81.1\% & - & - & - & - \\
    & FlowFormer~\cite{flow:huang_flowformer_ECCV_2022} & 3.02 & 82.6\% & - & - & - & - \\ 
    \midrule
    \multirow{3}*{PC} 
    & Meteornet*~\cite{liu2019meteornet} & - & - & - & - & 0.209 & - \\ 
    & PointPWC~\cite{scene:wu_pointpwc_eccv_2020} & - & - & 0.112 & 51.8\% & - & - \\
    & SCTN~\cite{scene:li_sctn_AAAI_2022} & - & - & 0.038 & 84.7\% & - & - \\ 
    \midrule
    RGB+Depth
    & RAFT-3D~\cite{scene:teed_raft3d_cvpr_2021} & 2.37 & \textbf{87.1}\% & 0.062 & 84.5\% & 0.089 & 71.1\% \\ 
    \midrule
    \multirow{2}*{RGB+PC}
    & DeepLiDARFlow~\cite{scene:rishav_deeplidarflow_iros_2020} & 6.04 & 47.1\% & - & 27.2\% & - & - \\
    & CamLiFlow~\cite{scene:liu_camliflow_cvpr_2022} & 2.20 & 84.6\% & 0.033 & 91.7\% & 0.059 & 86.0\% \\
    \midrule
    RGB+Event
    & RAFT+Event & 2.49 & 84.6\% & - & - & - & - \\
    \midrule
    \multirow{2}*{RGB+PC+Event}
    & CamLiFlow+Event & 1.56 & 84.4\% & 0.028 & 91.7\% & 0.048 & 85.9\% \\
    & RPEFlow~(Ours) & \textbf{1.40} & 86.2\% & \textbf{0.024} & \textbf{93.1}\% & \textbf{0.042} & \textbf{88.0}\% \\
    \bottomrule
    \end{tabular}
    }

    \label{tab:things}
\end{table*}

\section{Experiment}

\subsection{Implementation Details}
\noindent \textbf{Datasets.} 
Since there is no large-scale scene flow dataset with real event data, we use synthetic data for pretraining. 
Follow the preprocess pipeline~\cite{scene:liu_flownet3d_cvpr_2019, scene:liu_camliflow_cvpr_2022}, we generate point clouds from depth images for FlyingThings3D~\cite{flowdatasets:Mayer_FlyingThing3d_CVPR_2016} dataset, which contains 19,640 and 3,824 RGB-PointColud pairs for ``train'' and ``val'' splits. 
We use the popular video-to-events conversion method~\cite{eventdatasets:Gehrig_VideoToEvent_CVPR_2020} to generate the corresponding events. 
In addition, we use kubric~\cite{greff2022kubric} to simulate 15,367 RGB-PointCloud-Event pairs with rich annotations (including optical flow and scene flow ground truths), denoted as \textbf{EKubric}, which aims to simulate photo-realistic scenes with collision detection, gravity model and ambient illumination and has more object kinds than FlyingThings3D. 
We also use the DSEC~\cite{eventflow:Gehrig_DenseRAFTFlow_3DV_2021} dataset, which contains 8,170 pairs of real-captured samples in driving scenarios. 

\begin{table*}[tbp]
    \caption{\textbf{Finetuned performance comparison} on the test split of our simulated EKubric dataset.} 
    \renewcommand\arraystretch{0.75}
    \centering
    \setlength{\tabcolsep}{2.8mm}{
    \small
    \begin{tabular}{cccccccc}
    \toprule
    Input & Method & $\rm {EPE}_{2D}$ & $\rm {ACC}_{1px}$ & $\rm {EPE}_{3D}^{N.Occ}$ & $\rm {ACC}_{.05}^{N.Occ}$ & $\rm {EPE}_{3D}^{Full}$ & $\rm {ACC}_{.05}^{Full}$ \\
    \midrule
    \multirow{2}*{RGB} 
    & RAFT~\cite{flow:Teed_RAFT_ECCV_2020} & 0.757 & 93.70\% & - & - & - & - \\
    & FlowFormer~\cite{flow:huang_flowformer_ECCV_2022} & 0.683 & 93.92\% & - & - & - & - \\ 
    \midrule
    {RGB+Depth}
    & RAFT-3D~\cite{scene:teed_raft3d_cvpr_2021} & 0.715 & 94.33\% & 0.016 & 95.20\% & 0.049 & 92.62\% \\
    \midrule
    {RGB+PC}
    & CamLiFlow~\cite{scene:liu_camliflow_cvpr_2022} & 0.761 & 95.00\% & 0.009 & 98.39\% & 0.032 & 94.90\% \\
    \midrule
    {RGB+Event}
    & RAFT+Event & 0.487 & 95.25\% & - & - & - & - \\
    \midrule
    \multirow{2}*{RGB+PC+Event}
    & CamLiFlow+Event & 0.505 & 95.41\% & 0.008 & 98.48\% & 0.031 & 95.01\% \\
    & RPEFlow~(Ours) & \textbf{0.442} & \textbf{96.08}\% & \textbf{0.007} & \textbf{98.68}\% & \textbf{0.027} & \textbf{95.30}\% \\
    \bottomrule
    \end{tabular}
    }
    \label{tab:finetune_ekubric}
\end{table*}

\begin{figure*}[tbp]
\small
\centering

\begin{minipage}[t]{0.162\linewidth}
\centering
\includegraphics[width=\textwidth]{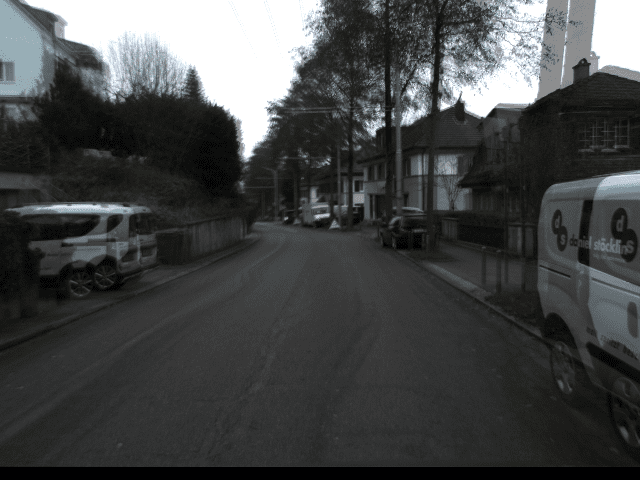} \\
\vspace{0.01cm} 
\includegraphics[width=\textwidth]{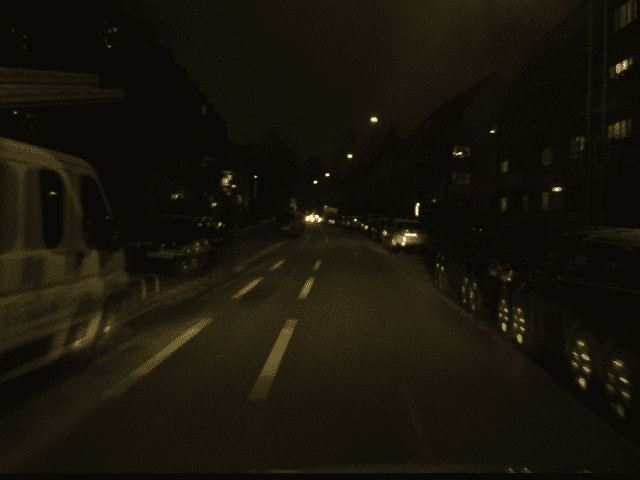} \\
Overlapped Frames
\end{minipage}%
\hspace{0.001\linewidth}
\begin{minipage}[t]{0.162\linewidth}
\centering
\includegraphics[width=0.98\textwidth, frame]{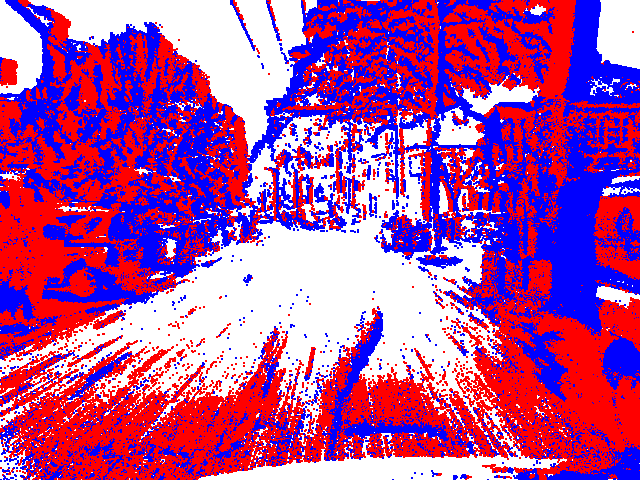} \\
\vspace{0.01cm} 
\includegraphics[width=0.98\textwidth, frame]{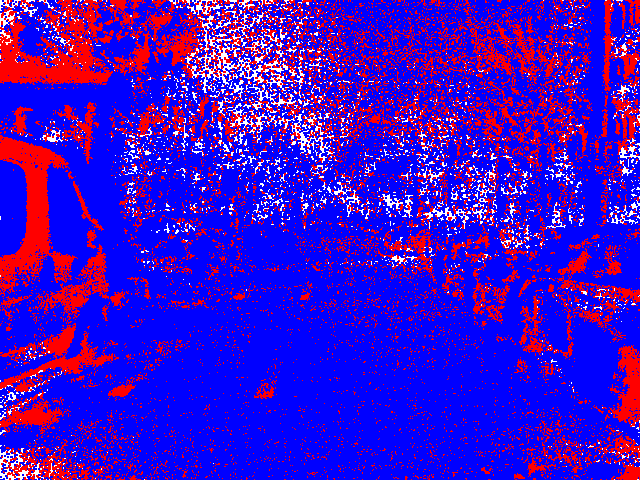} \\
Events
\end{minipage}%
\hspace{0.001\linewidth}
\begin{minipage}[t]{0.162\linewidth}
\centering
\includegraphics[width=0.98\textwidth, frame]{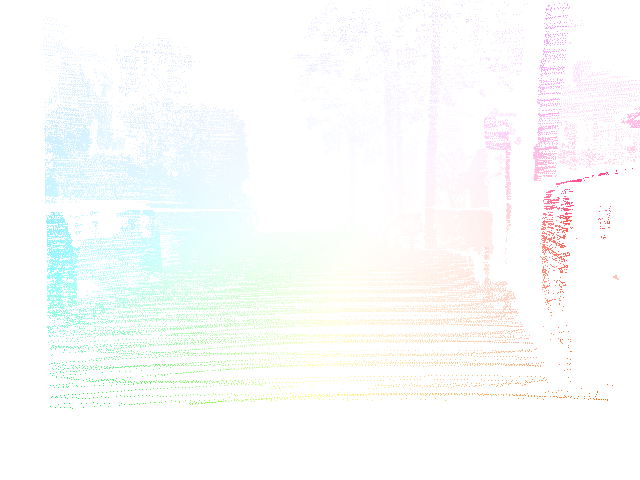} \\
\vspace{0.01cm} 
\includegraphics[width=0.98\textwidth, frame]{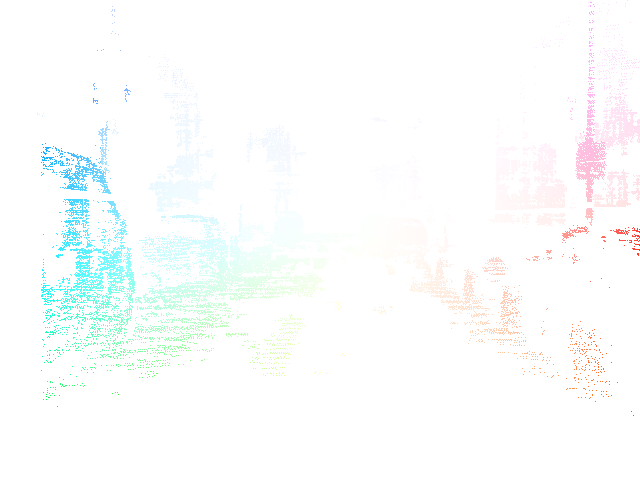} \\
Flow GT
\end{minipage}%
\hspace{0.001\linewidth}
\begin{minipage}[t]{0.162\linewidth}
\centering
\includegraphics[width=\textwidth]{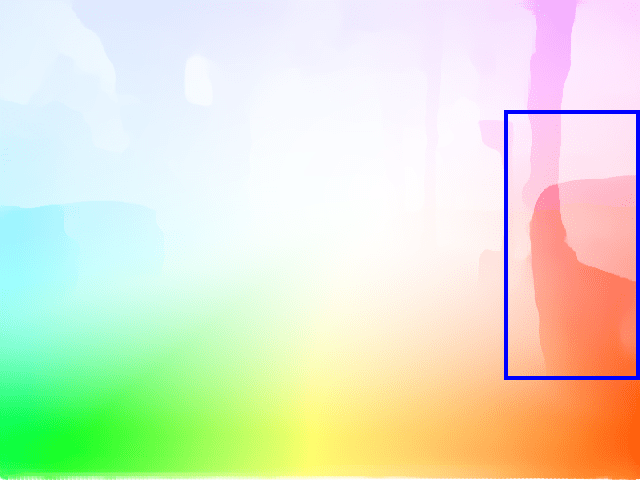} \\
\vspace{0.01cm} 
\includegraphics[width=\textwidth]{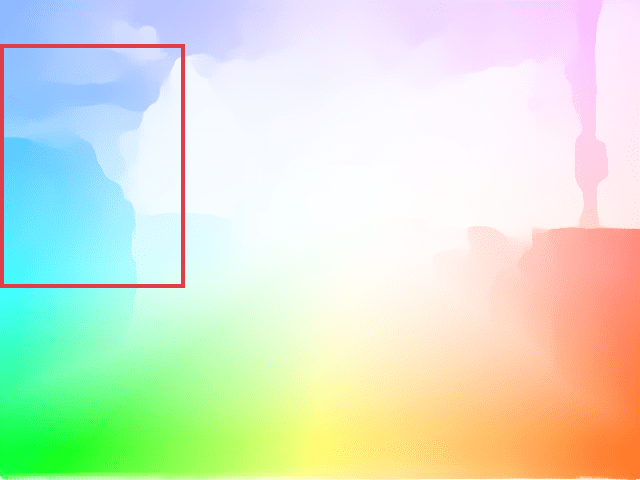} \\
CamLiFlow~\cite{scene:liu_camliflow_cvpr_2022}
\end{minipage}%
\hspace{0.001\linewidth}
\begin{minipage}[t]{0.162\linewidth}
\centering
\includegraphics[width=\textwidth]{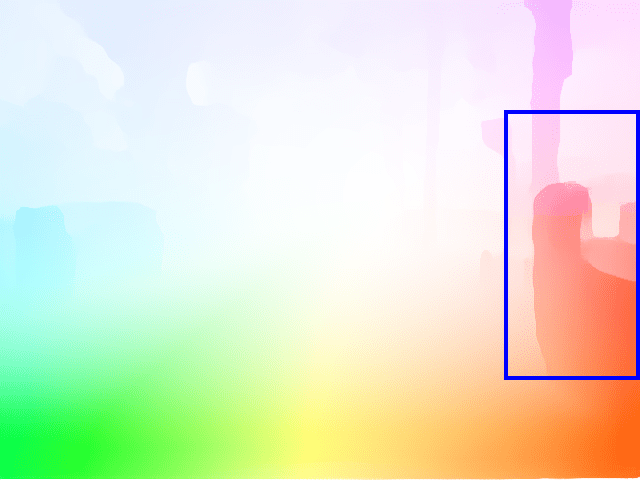} \\
\vspace{0.01cm} 
\includegraphics[width=\textwidth]{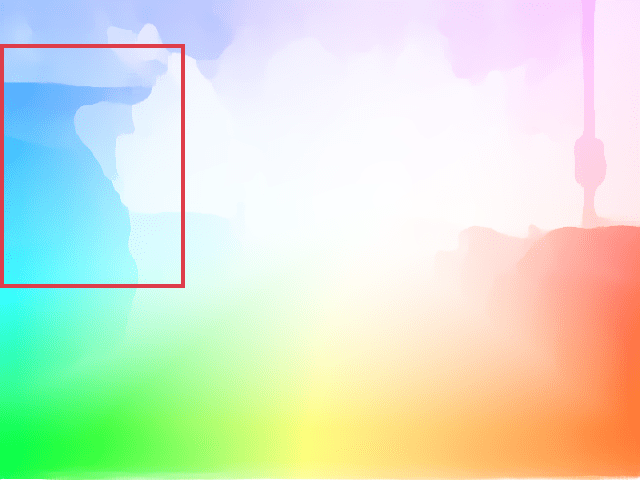} \\
\vspace{-0.05cm}
CamLiFlow+Events
\end{minipage}%
\hspace{0.001\linewidth}
\begin{minipage}[t]{0.162\linewidth}
\centering
\includegraphics[width=\textwidth]{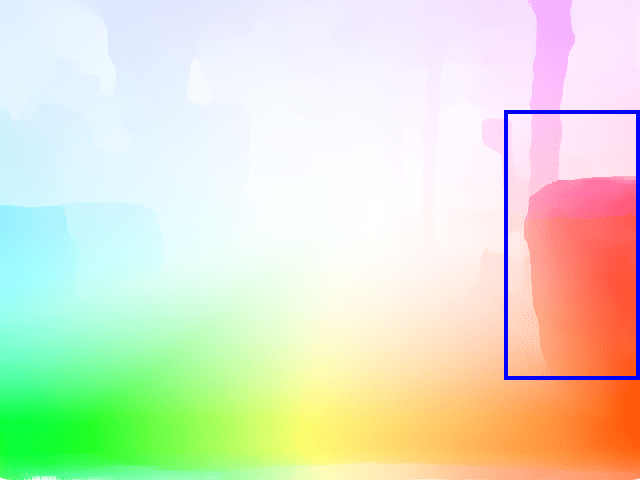} \\
\vspace{0.01cm} 
\includegraphics[width=\textwidth]{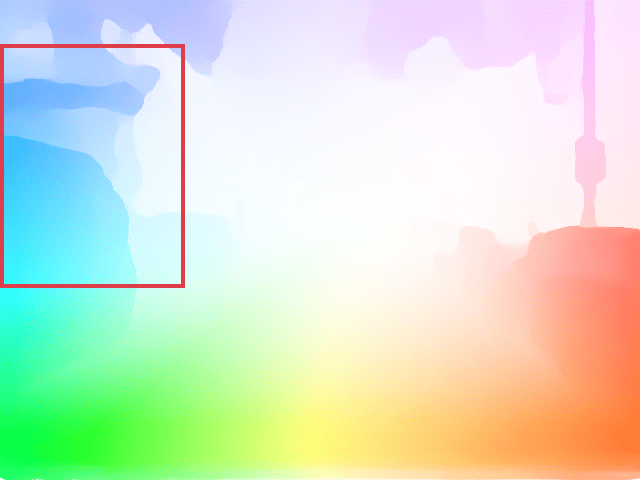} \\
\vspace{-0.05cm}
RPEFlow~(Ours)
\end{minipage}%
\\

\begin{minipage}[t]{0.185\linewidth}
\centering
\includegraphics[width=0.96\textwidth, frame]{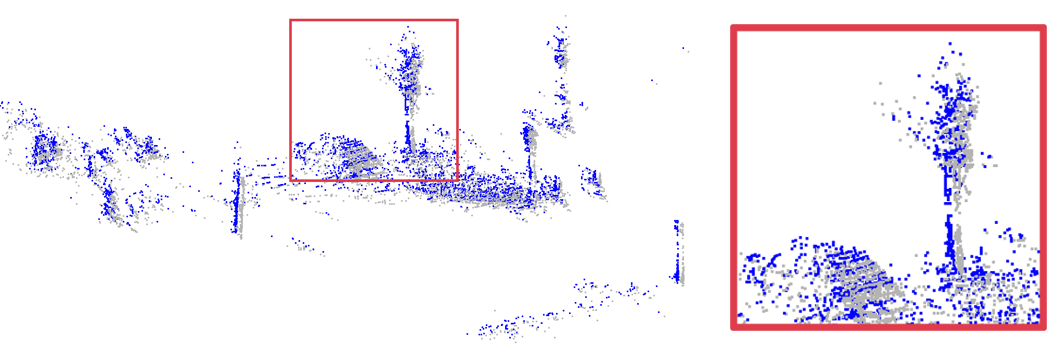} \\
\vspace{0.01cm} 
\includegraphics[width=0.96\textwidth, frame]{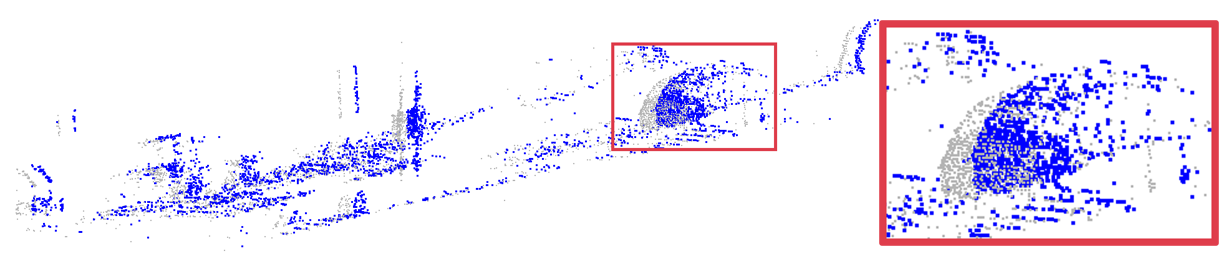} \\
\vspace{-0.05cm}
Point Clouds
\end{minipage}%
\hspace{0.005\linewidth}
\begin{minipage}[t]{0.185\linewidth}
\centering
\includegraphics[width=0.96\textwidth, frame]{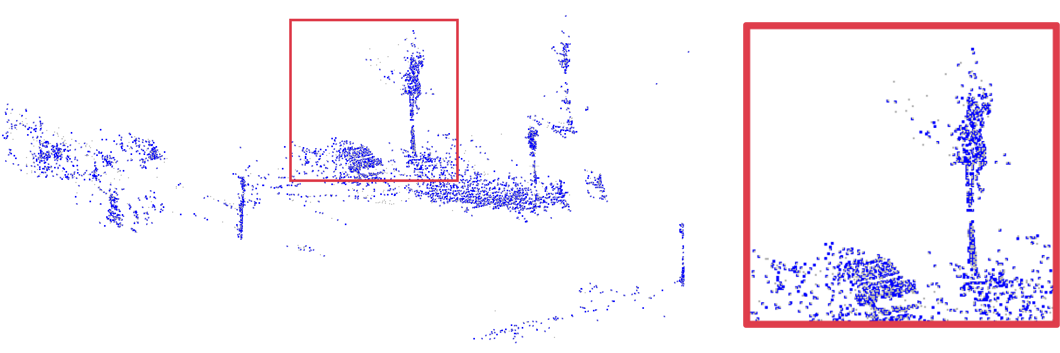} \\
\vspace{0.01cm} 
\includegraphics[width=0.96\textwidth, frame]{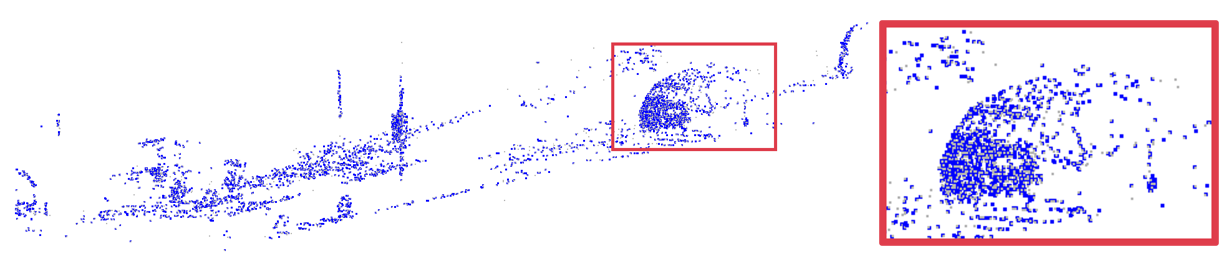} \\
\vspace{-0.05cm}
Scene Flow GT
\end{minipage}%
\hspace{0.005\linewidth}
\begin{minipage}[t]{0.185\linewidth}
\centering
\includegraphics[width=0.96\textwidth, frame]{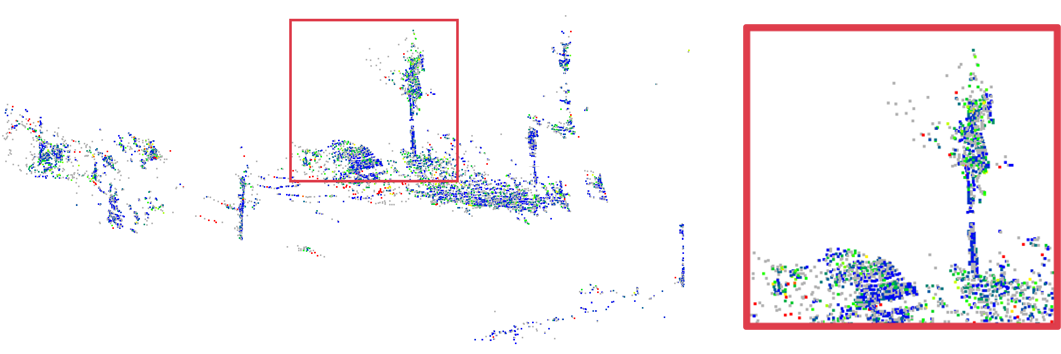} \\
\vspace{0.01cm} 
\includegraphics[width=0.96\textwidth, frame]{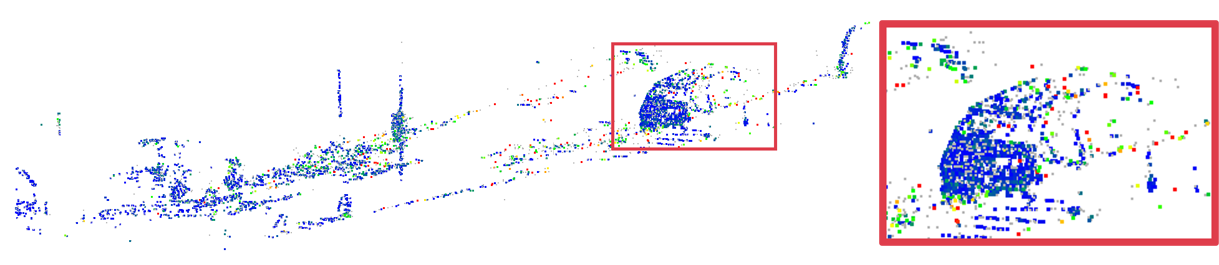} \\
\vspace{-0.05cm}
CamLiFlow~\cite{scene:liu_camliflow_cvpr_2022}
\end{minipage}%
\hspace{0.005\linewidth}
\begin{minipage}[t]{0.185\linewidth}
\centering
\includegraphics[width=0.96\textwidth, frame]{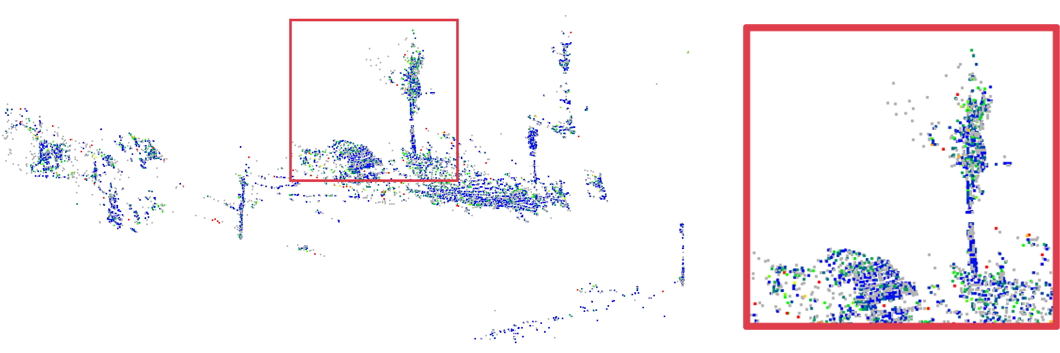} \\
\vspace{0.01cm} 
\includegraphics[width=0.96\textwidth, frame]{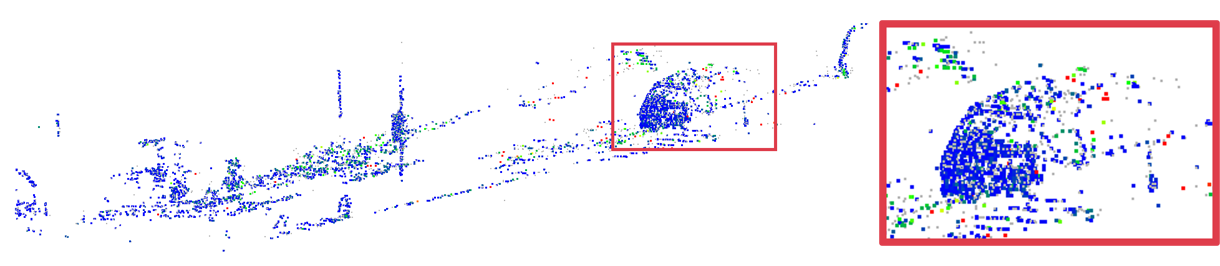} \\
\vspace{-0.05cm}
CamLiFlow+Events
\end{minipage}%
\hspace{0.005\linewidth}
\begin{minipage}[t]{0.185\linewidth}
\centering
\includegraphics[width=0.96\textwidth, frame]{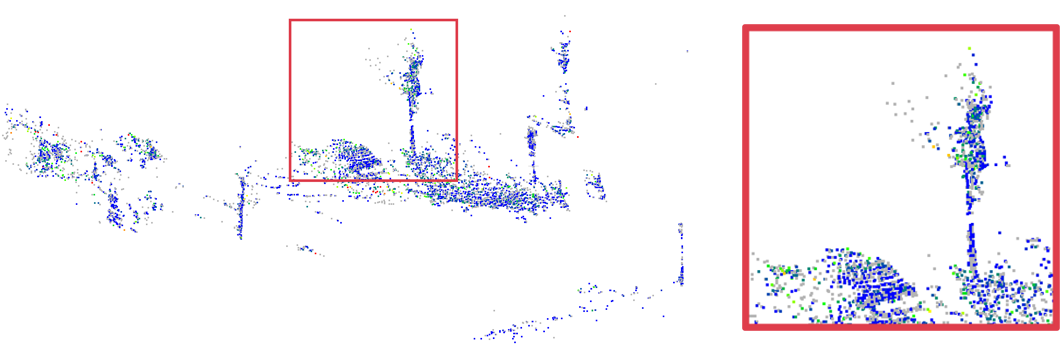} \\
\vspace{0.01cm} 
\includegraphics[width=0.96\textwidth, frame]{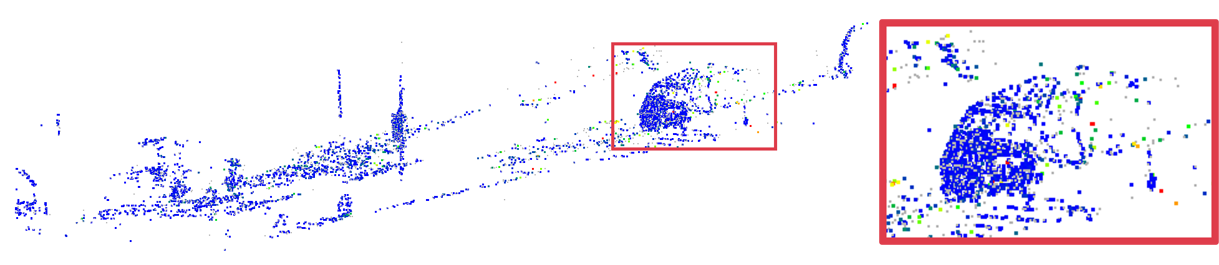} \\
\vspace{-0.05cm}
RPEFlow~(Ours)
\end{minipage}%
\begin{minipage}[t]{0.001\linewidth}
\centering
\raisebox{-0.42\height}{\includegraphics[height=1.7cm]{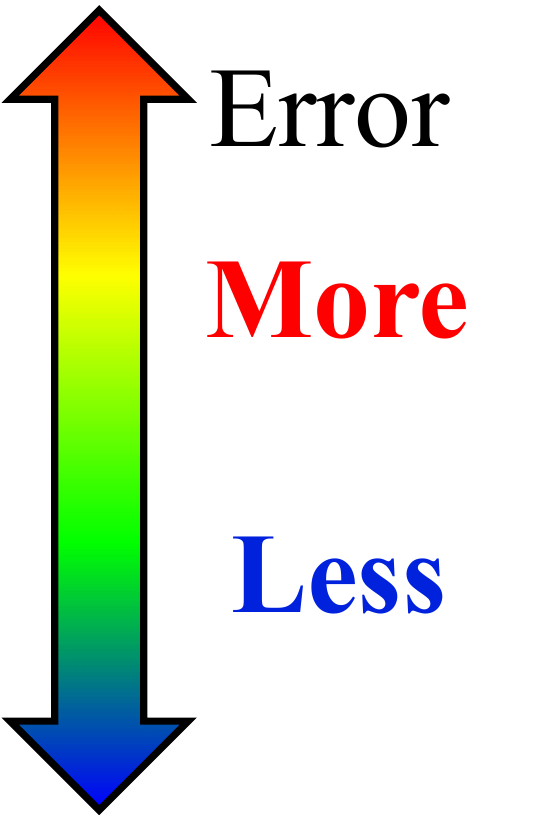}}
\end{minipage}%

\centering
\caption{\textbf{Visual comparisons} on real-captured data, \ie the ``val'' split of DSEC~\cite{eventflow:Gehrig_DenseRAFTFlow_3DV_2021} dataset. 
} 
\label{viz:dsec}
\end{figure*}

\noindent \textbf{Training, Hyper-parameters and Metrics.}
Our model is trained with PyTorch on four RTX3090 GPUs and evaluated on one.
We use the Adam optimizer with weight decay $10^{-6}$. 
The number of event bins is $B\!=\!10$, pyramid layers is $L\!=\!5$. 
Loss weights are $\alpha\!=\!10.0$, $\beta\!=\!0.01$, $\lambda_l\!=\!2^{(l-2)}$ for $l \!\in\! [1, L]$. 
Following~\cite{scene:teed_raft3d_cvpr_2021, scene:liu_camliflow_cvpr_2022}, we evaluate using 2D and 3D end-point error ({\small $\rm {EPE}_{2D}$} and {\small $\rm {EPE}_{3D}$}), and {\small $\rm {ACC}_{1px}$} and {\small $\rm {ACC}_{.05}$} to measure the portion of accuracy within 1 pixel and 5cm. 
The scene flow metrics with {\small $\rm N.Occ$} superscript indicates only the not occluded positions are calculated, while {\small $\rm Full$} or none indicates all positions including occlusion. 
More details about datasets and training are provided in the supplementary materials.

\subsection{Comparisons with Synthetic Events}

Following the conventional setting, our model is first pre-trained on FlyingThings3D~\cite{flowdatasets:Mayer_FlyingThing3d_CVPR_2016}, and Table~\ref{tab:things} shows the quantitative comparison performance on the ``val'' split. 
Both results are evaluated using their pre-trained model on FlyingThings3D. Meteornet*~\cite{liu2019meteornet} has not released model for scene flow estimation, so we compare the EPE3D result from its paper.
We also finetune them on our simulated EKubric dataset, and the performance comparisons on the test set are shown in Table~\ref{tab:finetune_ekubric}, where the compared models were all pre-trained on FlyingThings3D and then fine-tuned on EKubric with the same training settings. 
Note that we did not evaluate the PC-only methods on our EKubric dataset due to their different data preprocessing strategies.

We report the results of two representative methods~\cite{flow:Teed_RAFT_ECCV_2020, scene:wu_pointpwc_eccv_2020} and two latest methods~\cite{flow:huang_flowformer_ECCV_2022, scene:li_sctn_AAAI_2022} using only one modal for one task, and find that combining multiple modalities can significantly improve model accuracy for both optical flow and scene flow estimation. 
Furthermore, the methods that combine all three modalities, \ie~RGB+PC+Event, achieve better results than the rest~\cite{scene:teed_raft3d_cvpr_2021, scene:rishav_deeplidarflow_iros_2020, scene:liu_camliflow_cvpr_2022}, which illustrates the significant benefit of introducing event data for accurate motion estimation. 
With the same input setting, we compare our model with CamLiFlow+Event and observe improved accuracy, which is because our proposed multimodal attention fusion module is able to fully mine valuable information from continuous event data for accurate motion estimation. 
The comparison of visualization results in Fig.~\ref{viz:things_kubric} and in the supplementary materials is also consistent with the above observations, especially in high dynamic and detail moving or motion-blurred areas. 

\begin{table}[tbp]
    \caption{\textbf{Finetuned Performance comparison} on the ``val'' split of DSEC~\cite{eventflow:Gehrig_DenseRAFTFlow_3DV_2021} dataset.}
    \renewcommand\arraystretch{0.75}
    \centering
    \begin{threeparttable}
    \setlength{\tabcolsep}{1mm}{
    \small
    \begin{tabular}{ccccc}
    \toprule
    Method & $\rm {EPE}_{2D}$ & $\rm {ACC}_{1px}$ & $\rm {EPE}_{3D}^{Full}$ & $\rm {ACC}_{.05}^{Full}$ \\
    \midrule
    RAFT~\cite{flow:Teed_RAFT_ECCV_2020} & 0.572 & 89.63\% & - & - \\
    E-RAFT*~\cite{eventflow:Gehrig_DenseRAFTFlow_3DV_2021} & 0.473 & 92.11\% & - & - \\
    \midrule
    RAFT-3D~\cite{scene:teed_raft3d_cvpr_2021} & 0.567 & 90.55\% & 0.140 & 51.30\% \\
    CamLiFlow~\cite{scene:liu_camliflow_cvpr_2022} & 0.383 & 94.92\% & 0.120 & 53.49\% \\
    \midrule
    RAFT+Event & 0.537 & 90.08\% & - & - \\
    CamLiFlow+Event & 0.361 & 95.07\% & 0.116 & 55.26\% \\
    \midrule
    RPEFlow~(Ours) & \textbf{0.332} & \textbf{95.27}\% & \textbf{0.104} & \textbf{60.50}\% \\
    \bottomrule
    \end{tabular}
    }
    \end{threeparttable}
    \label{tab:dsec}
\end{table}

\subsection{Comparisons with Real Events}
We further conduct experiments on the real-captured DSEC~\cite{eventflow:Gehrig_DenseRAFTFlow_3DV_2021} dataset. 
We divide the public set into \enquote{train} and \enquote{val} splits for finetuning and evaluation. 
Since the dense depth required by RAFT-3D~\cite{scene:teed_raft3d_cvpr_2021} is not available, we use CFNet~\cite{shen2021cfnet} to obtain a pseudo-dense depth map by stereo matching as input.
E-RAFT*~\cite{eventflow:Gehrig_DenseRAFTFlow_3DV_2021} does not have a publicly available training code, thus we can only use the pretrained model on the entire DSEC dataset including the ``val'' split (marked with *).
Both quantitative and qualitative (Table~\ref{tab:dsec} and Fig.~\ref{viz:dsec}) comparisons demonstrate the superiority of our model for real-captured data, which is consistent with the observations on synthetic datasets above. 
In particular, our method performs better in night driving (the 2nd sample in Fig.~\ref{viz:dsec}), because the event camera is still sensitive to brightness changes even in low-light scenes. 
These comparisons further illustrate the importance of introducing events and the applicability of our model to practical needs.

\begin{table}[tbp]
\caption{\textbf{Ablation Studies.} The results validate the effectiveness for introducing event data, attention fusion and mutual information regularization, respectively. 
}
\renewcommand\arraystretch{0.75}
\centering
\resizebox{\linewidth}{!}{
\setlength{\tabcolsep}{1pt}{
    \footnotesize
    \begin{tabular}{cccccccc}
    \toprule
    & Event & Fusion & MI & $\rm {EPE}_{2D}$ & $\rm {ACC}_{1px}$ & $\rm {EPE}_{3D}^{Full}$ & $\rm {ACC}_{.05}^{Full}$ \\
    \midrule
    (a) & - & Concat & - & 2.200 & 84.62\% & 0.059 & 86.02\% \\
    (b) & - & Attention & - & 2.133 & 84.74\% & 0.058 & 86.53\% \\
    (c) & - & Attention & \checkmark & 2.067 & 84.92\% & 0.055 & 86.85\% \\
    (d) & \checkmark & Concat\textsubscript{W/o E}  & - & 1.561 & 84.37\% & 0.048 & 85.86\%\\
    (e) & \checkmark & Concat\textsubscript{W/ E} & - & 1.519 & 85.34\% & 0.046 & 86.63\% \\
    (f) & \checkmark & Attention & - & 1.494 & 86.01\% & 0.043 & 87.39\% \\
    \midrule
    (g) & \checkmark & Attention & \checkmark & \textbf{1.402} & \textbf{86.22}\% & \textbf{0.042} & \textbf{88.01}\% \\
    \bottomrule
    \end{tabular}
}
}
\label{tab:ablation}
\end{table}

\subsection{Ablation Studies}
In Table~\ref{tab:ablation}, We conduct ablation experiments to verify the contribution of each component in our model. 
All variations are trained on the \enquote{training} split and evaluated on the \enquote{val} split of FlyingThings3D~\cite{flowdatasets:Mayer_FlyingThing3d_CVPR_2016} dataset. 
In addition to the following discussion, in the supplementary materials we analyze the impact of real and simulated events and the role of combining the two tasks. 

\noindent \textbf{Event data.} 
As we are the first to introduce the event data for joint optical flow and scene flow estimation, we explore the effectiveness of event data in two aspects. 
Firstly, we remove the event data in our framework (see Table~\ref{tab:ablation}~(a)-(d) and (c)-(g)), leading to significant performance degradation, especially in optical flow error. 
This is in line with our claim that event data with continuous observations of scene brightness changes can provide significant help for accurate motion estimation. 
Secondly, in order to validate the benefit of introducing events to other methods, 
we concatenate the extracted event feature as an additional input of flow decoder for RAFT~\cite{flow:Teed_RAFT_ECCV_2020} and CamLiFlow~\cite{scene:liu_camliflow_cvpr_2022}, denoted as RAFT+Event and CamLiFlow+Event, respectively. 
Results in Table~\ref{tab:things}, \ref{tab:finetune_ekubric} and \ref{tab:dsec} show that introducing event data to existing models can also significantly improve the performance, rather than just on our proposed framework. 

\noindent \textbf{Fusion Structure.}
Unlike CamLiFlow~\cite{scene:liu_camliflow_cvpr_2022} that concat the features from two modalities,
we propose an attention-based multimodal fusion module. 
We conduct experiments with both no-event and with event settings (see Table~\ref{tab:ablation}~(a)-(b) and (d)(e)-(f)), and our proposed attention fusion improves both 2D and 3D motion estimation performance. 
This shows that our multimodal attention fusion module can effectively explore the complementary information and correlations among the three modalities and generate fused features that are suitable for subsequent motion estimation. 

\noindent \textbf{Mutual Information.}
The purpose of the mutual information regularization term is to explicitly constrain the network to learn complementary information from different modalities of data. 
To verify its contribution to motion estimation, we conduct experiments with both no-event and with event settings (see Table~\ref{tab:ablation}~(b)-(c) and (f)-(g)). 
The results show that with this explicit constraint, the fused features can further improve the accuracy of motion estimation. 
In addition, we believe that the multimodal attention fusion module and mutual information regularization term both play an important role in motion estimation, because their contribution to 2D and 3D motion estimation is significant in both the with-event and without-event settings.

\subsection{Computation Cost}
Our model is relatively efficient compared to the latest unimodal and multimodal methods in model size and run-time, i.e., FlowFormer~\cite{flow:huang_flowformer_ECCV_2022}: 17.6M, 1.35s, SCTN~\cite{scene:li_sctn_AAAI_2022}: 7.8M, 242ms, RAFT-3D~\cite{scene:teed_raft3d_cvpr_2021}: 45M, 593ms, CamLiFlow~\cite{scene:liu_camliflow_cvpr_2022}: 7.7M, 88ms ($\approx$2s with refine) and Ours: 9.75M, 112ms (both on a single RTX3090 GPU with 1280$\times$720 size). 

\begin{table}[tbp]
\renewcommand\arraystretch{0.75}
\centering
\resizebox{\linewidth}{!}{
\setlength{\tabcolsep}{1pt}{
    \small
    \begin{tabular}{ccccccc}
    \toprule
    Setting & $\rm {EPE}_{2D}$ & $\rm {ACC}_{1px}$ & $\rm {EPE}_{3D}^{N.Occ}$ & $\rm {ACC}_{.05}^{N.Occ}$ & $\rm {EPE}_{3D}^{Full}$ & $\rm {ACC}_{.05}^{Full}$ \\
    \midrule
    2D only & 1.937 & 83.17\% & - & - & - & - \\
    3D only & - & - & 0.043 & 86.37\% & 0.078 & 79.05\% \\
    \midrule
    2D\&3D & \textbf{1.402} & \textbf{86.22}\% & \textbf{0.024} & \textbf{93.14}\% & \textbf{0.042} & \textbf{88.01}\% \\
    \bottomrule
    \end{tabular}
}
}
\caption{\textbf{Joint vs independent tasks.} These models both use three modalities and the results validate that combining the two tasks makes better use of motion information. }
\label{tab:joint}
\end{table}

\subsection{Joint 2D and 3D estimation.}
In previous comparisons, the methods of joint optical flow and scene flow estimation perform significantly better than independent methods. 
To further verify the necessity of jointing these two tasks, we compare the models that only supervise the optical flow or scene flow estimation in Table~\ref{tab:joint}. 
Although using all three modalities for a single task (2D only and 3D only) is more beneficial than the methods using fewer modality data in Table~1 of main paper. 
However, combining the two tasks in 2D and 3D benefits from the tight correlation between the two deeply supervised branches, allowing more adequate exploiting of the correlation between 2D and 3D input modalities and more accurate estimating of 2D and 3D motion jointly.

\section{Conclusion}
We introduce a multimodal fusion framework for joint 2D optical flow and 3D scene flow estimation by fusing RGB images, point clouds and event data. 
Our contributions are threefold: \textbf{1)} By incorporating event data, our new framework could handle highly dynamic scenes. 
\textbf{2)} We fuse representations of the three very different modalities both implicitly and explicitly through multimodal attention and mutual information regularization, respectively. 
\textbf{3)} We contribute a new simulation dataset to further advocate research in this direction. 
Our work shows that event cameras can play an important role in 2D and 3D motion estimation, and reveals the prospect of event-based 3D vision. 

\noindent\textbf{Limitation.}
Our model is not specially designed for extreme situations such as dark nights or sensor failures, and we plan to address them in the future. 

\section{Acknowledgments}
This research was supported in part by National Natural Science Foundation of China (62271410), the Fundamental Research Funds for the Central Universities, and Zhejiang Lab (NO.2021MC0AB05). Zhexiong Wan is sponsored by the scholarship from China Scholarship Council (202306290193) and the Innovation Foundation for Doctor Dissertation of Northwestern Polytechnical University (CX2023013). 


{\small
\bibliographystyle{unsrt}
\bibliography{EventSceneFlow_Reference}

\begin{thebibliography}{10}

\bibitem{flow:Sun_PWCNet_CVPR_2018}
Deqing Sun, Xiaodong Yang, Ming-Yu Liu, and Jan Kautz.
\newblock Pwc-net: Cnns for optical flow using pyramid, warping, and cost
  volume.
\newblock In {\em IEEE Conference on Computer Vision and Pattern Recognition
  (CVPR)}, pages 8934--8943, 2018.

\bibitem{flow:Teed_RAFT_ECCV_2020}
Zachary Teed and Jia Deng.
\newblock Raft: Recurrent all-pairs field transforms for optical flow.
\newblock In {\em European Conference on Computer Vision (ECCV)}, pages
  402--419, 2020.

\bibitem{flow:huang_flowformer_ECCV_2022}
Zhaoyang Huang, Xiaoyu Shi, Chao Zhang, Qiang Wang, Ka~Chun Cheung, Hongwei
  Qin, Jifeng Dai, and Hongsheng Li.
\newblock {FlowFormer}: A transformer architecture for optical flow.
\newblock In {\em European Conference on Computer Vision (ECCV)}, pages
  668--685, 2022.

\bibitem{scene:hur_selfscene_CVPR_2020}
Junhwa Hur and Stefan Roth.
\newblock Self-supervised monocular scene flow estimation.
\newblock In {\em IEEE Conference on Computer Vision and Pattern Recognition
  (CVPR)}, pages 7396--7405, 2020.

\bibitem{scene:guizilini_DRAFT_RAL_2022}
Vitor Guizilini, Kuan-Hui Lee, Rare{\c{s}} Ambru{\c{s}}, and Adrien Gaidon.
\newblock Learning optical flow, depth, and scene flow without real-world
  labels.
\newblock {\em IEEE Robotics and Automation Letters}, 7(2):3491--3498, 2022.

\bibitem{scene:ilg_occlusions_ECCV_2018}
Eddy Ilg, Tonmoy Saikia, Margret Keuper, and Thomas Brox.
\newblock Occlusions, motion and depth boundaries with a generic network for
  disparity, optical flow or scene flow estimation.
\newblock In {\em European Conference on Computer Vision (ECCV)}, pages
  614--630, 2018.

\bibitem{scene:ma_deeprigid_CVPR_2019}
Wei-Chiu Ma, Shenlong Wang, Rui Hu, Yuwen Xiong, and Raquel Urtasun.
\newblock Deep rigid instance scene flow.
\newblock In {\em IEEE Conference on Computer Vision and Pattern Recognition
  (CVPR)}, pages 3614--3622, 2019.

\bibitem{scene:wu_pointpwc_eccv_2020}
Wenxuan Wu, Zhi~Yuan Wang, Zhuwen Li, Wei Liu, and Li~Fuxin.
\newblock Pointpwc-net: Cost volume on point clouds for (self-) supervised
  scene flow estimation.
\newblock In {\em European Conference on Computer Vision (ECCV)}, pages
  88--107. Springer, 2020.

\bibitem{scene:li_sctn_AAAI_2022}
Bing Li, Cheng Zheng, Silvio Giancola, and Bernard Ghanem.
\newblock {SCTN}: Sparse convolution-transformer network for scene flow
  estimation.
\newblock In {\em AAAI Conference on Artificial Intelligence (AAAI)},
  volume~36, pages 1254--1262, 2022.

\bibitem{scene:teed_raft3d_cvpr_2021}
Zachary Teed and Jia Deng.
\newblock {RAFT-3D}: Scene flow using rigid-motion embeddings.
\newblock In {\em IEEE Conference on Computer Vision and Pattern Recognition
  (CVPR)}, pages 8375--8384, 2021.

\bibitem{scene:yang_flowexpan_CVPR_2020}
Gengshan Yang and Deva Ramanan.
\newblock Upgrading optical flow to 3d scene flow through optical expansion.
\newblock In {\em IEEE Conference on Computer Vision and Pattern Recognition
  (CVPR)}, pages 1334--1343, 2020.

\bibitem{scene:rishav_deeplidarflow_iros_2020}
Rishav Rishav, Ramy Battrawy, Ren{\'e} Schuster, Oliver Wasenm{\"u}ller, and
  Didier Stricker.
\newblock Deeplidarflow: A deep learning architecture for scene flow estimation
  using monocular camera and sparse lidar.
\newblock In {\em IEEE/RJS International Conference on Intelligent Robots and
  Systems (IROS)}, pages 10460--10467. IEEE, 2020.

\bibitem{scene:liu_camliflow_cvpr_2022}
Haisong Liu, Tao Lu, Yihui Xu, Jia Liu, Wenjie Li, and Lijun Chen.
\newblock Camliflow: Bidirectional camera-lidar fusion for joint optical flow
  and scene flow estimation.
\newblock In {\em IEEE Conference on Computer Vision and Pattern Recognition
  (CVPR)}, pages 5791--5801, 2022.

\bibitem{flowdatasets:Geiger_kitti2012_cvpr_2012}
Andreas Geiger, Philip Lenz, and Raquel Urtasun.
\newblock Are we ready for autonomous driving? the kitti vision benchmark
  suite.
\newblock In {\em IEEE Conference on Computer Vision and Pattern Recognition
  (CVPR)}, pages 3354--3361, 2012.

\bibitem{flowdatasets:menze_object_kitti_2015}
Moritz Menze and Andreas Geiger.
\newblock Object scene flow for autonomous vehicles.
\newblock In {\em IEEE Conference on Computer Vision and Pattern Recognition
  (CVPR)}, pages 3061--3070, 2015.

\bibitem{porzi_trackseg_auto_CVPR_2020}
Lorenzo Porzi, Markus Hofinger, Idoia Ruiz, Joan Serrat, Samuel~Rota Bulo, and
  Peter Kontschieder.
\newblock Learning multi-object tracking and segmentation from automatic
  annotations.
\newblock In {\em IEEE Conference on Computer Vision and Pattern Recognition
  (CVPR)}, pages 6846--6855, 2020.

\bibitem{wang_multiple_correlation_CVPR_2021}
Qiang Wang, Yun Zheng, Pan Pan, and Yinghui Xu.
\newblock Multiple object tracking with correlation learning.
\newblock In {\em IEEE Conference on Computer Vision and Pattern Recognition
  (CVPR)}, pages 3876--3886, 2021.

\bibitem{zhang_flowfusion_ICRA_2020}
Tianwei Zhang, Huayan Zhang, Yang Li, Yoshihiko Nakamura, and Lei Zhang.
\newblock Flowfusion: Dynamic dense rgb-d slam based on optical flow.
\newblock In {\em IEEE International Conference on Robotics and Automation
  (ICRA)}, pages 7322--7328. IEEE, 2020.

\bibitem{li_neuralscenefields_CVPR_2021}
Zhengqi Li, Simon Niklaus, Noah Snavely, and Oliver Wang.
\newblock Neural scene flow fields for space-time view synthesis of dynamic
  scenes.
\newblock In {\em IEEE Conference on Computer Vision and Pattern Recognition
  (CVPR)}, pages 6498--6508, 2021.

\bibitem{gallego_eventsurvey_TPAMI_2022}
Guillermo Gallego, Tobi Delbrück, Garrick Orchard, Chiara Bartolozzi, Brian
  Taba, Andrea Censi, Stefan Leutenegger, Andrew~J. Davison, Jörg Conradt,
  Kostas Daniilidis, and Davide Scaramuzza.
\newblock Event-based vision: A survey.
\newblock {\em IEEE Transactions on Pattern Analysis and Machine Intelligence
  (TPAMI)}, 44(1):154--180, 2022.

\bibitem{eventflow:Zhu_EVFlowNet_CVPR_2019}
Alex~Zihao Zhu, Liangzhe Yuan, Kenneth Chaney, and Kostas Daniilidis.
\newblock Unsupervised event-based learning of optical flow, depth, and
  egomotion.
\newblock In {\em IEEE Conference on Computer Vision and Pattern Recognition
  (CVPR)}, pages 989--997, 2019.

\bibitem{eventflow:Gehrig_DenseRAFTFlow_3DV_2021}
Mathias Gehrig, Mario Millh{\"a}usler, Daniel Gehrig, and Davide Scaramuzza.
\newblock E-raft: Dense optical flow from event cameras.
\newblock In {\em International Conference on 3D Vision (3DV)}, pages 197--206,
  2021.

\bibitem{eventflow:liu_adaptiveblock_BMVC_2018}
Min Liu and Tobias Delbr{\"u}ck.
\newblock Adaptive time-slice block-matching optical flow algorithm for dynamic
  vision sensors.
\newblock In {\em British Machine Vision Conference (BMVC)}, page 280, 2018.

\bibitem{eventflow:Pan_SingleImageFlow_CVPR_2020}
Liyuan Pan, Miaomiao Liu, and Richard Hartley.
\newblock Single image optical flow estimation with an event camera.
\newblock In {\em IEEE Conference on Computer Vision and Pattern Recognition
  (CVPR)}, pages 1669--1678, 2020.

\bibitem{eventflow:Wan_DCEIFlow_TIP_2022}
Zhexiong Wan, Yuchao Dai, and Yuxin Mao.
\newblock Learning dense and continuous optical flow from an event camera.
\newblock {\em IEEE Transactions on Image Processing (TIP)}, 31:7237--7251,
  2022.

\bibitem{eventdatasets:Zhu_MVSEC_RAL_2018}
Alex~Zihao Zhu, Dinesh Thakur, Tolga {\"O}zaslan, Bernd Pfrommer, Vijay Kumar,
  and Kostas Daniilidis.
\newblock The multivehicle stereo event camera dataset: An event camera dataset
  for 3d perception.
\newblock {\em IEEE Robotics and Automation Letters}, 3(3):2032--2039, 2018.

\bibitem{Isolating_Sources}
Ricky T.~Q. Chen, Xuechen Li, Roger~B Grosse, and David~K Duvenaud.
\newblock Isolating sources of disentanglement in variational autoencoders.
\newblock In {\em Advances in Neural Information Processing Systems (NeurIPS)},
  2018.

\bibitem{flowdatasets:Mayer_FlyingThing3d_CVPR_2016}
Nikolaus Mayer, Eddy Ilg, Philip Hausser, Philipp Fischer, Daniel Cremers,
  Alexey Dosovitskiy, and Thomas Brox.
\newblock A large dataset to train convolutional networks for disparity,
  optical flow, and scene flow estimation.
\newblock In {\em IEEE Conference on Computer Vision and Pattern Recognition
  (CVPR)}, pages 4040--4048, 2016.

\bibitem{flow:dosovitskiy_flownet_iccv_2015}
Alexey Dosovitskiy, Philipp Fischer, Eddy Ilg, Philip Hausser, Caner Hazirbas,
  Vladimir Golkov, Patrick Van Der~Smagt, Daniel Cremers, and Thomas Brox.
\newblock Flownet: Learning optical flow with convolutional networks.
\newblock In {\em IEEE International Conference on Computer Vision (ICCV)},
  pages 2758--2766, 2015.

\bibitem{flow:Ilg_Flownet2_cvpr_2017}
Eddy Ilg, Nikolaus Mayer, Tonmoy Saikia, Margret Keuper, Alexey Dosovitskiy,
  and Thomas Brox.
\newblock Flownet 2.0: Evolution of optical flow estimation with deep networks.
\newblock In {\em IEEE Conference on Computer Vision and Pattern Recognition
  (CVPR)}, pages 2462--2470, 2017.

\bibitem{flow:xu_gmflow_CVPR_2022}
Haofei Xu, Jing Zhang, Jianfei Cai, Hamid Rezatofighi, and Dacheng Tao.
\newblock Gmflow: Learning optical flow via global matching.
\newblock In {\em IEEE Conference on Computer Vision and Pattern Recognition
  (CVPR)}, pages 8121--8130, 2022.

\bibitem{scene:vedula_threed_TPAMI_2005}
Sundar Vedula, Peter Rander, Robert Collins, and Takeo Kanade.
\newblock Three-dimensional scene flow.
\newblock {\em IEEE Transactions on Pattern Analysis and Machine Intelligence
  (TPAMI)}, 27(3):475--480, 2005.

\bibitem{scene:huguet_variational_ICCV_2007}
Fr{\'e}d{\'e}ric Huguet and Fr{\'e}d{\'e}ric Devernay.
\newblock A variational method for scene flow estimation from stereo sequences.
\newblock In {\em IEEE International Conference on Computer Vision (ICCV)}.
  IEEE, 2007.

\bibitem{scene:brickwedde_monosf_ICCV_2019}
Fabian Brickwedde, Steffen Abraham, and Rudolf Mester.
\newblock Mono-sf: Multi-view geometry meets single-view depth for monocular
  scene flow estimation of dynamic traffic scenes.
\newblock In {\em IEEE International Conference on Computer Vision (ICCV)},
  pages 2780--2790, 2019.

\bibitem{scene:puy_flot_eccv_2020}
Gilles Puy, Alexandre Boulch, and Renaud Marlet.
\newblock Flot: Scene flow on point clouds guided by optimal transport.
\newblock In {\em European Conference on Computer Vision (ECCV)}, pages
  527--544. Springer, 2020.

\bibitem{eventflow:Gallego_Unifyingcontrastmax_cvpr_2018}
Guillermo Gallego, Henri Rebecq, and Davide Scaramuzza.
\newblock A unifying contrast maximization framework for event cameras, with
  applications to motion, depth, and optical flow estimation.
\newblock In {\em IEEE Conference on Computer Vision and Pattern Recognition
  (CVPR)}, pages 3867--3876, 2018.

\bibitem{eventflow:ieng_event3dflow4dsubspace_2017}
Sio-Hoi Ieng, Jo{\~a}o Carneiro, and Ryad~B Benosman.
\newblock Event-based 3d motion flow estimation using 4d spatio temporal
  subspaces properties.
\newblock {\em Frontiers in Neuroscience}, 10:596, 2017.

\bibitem{eventflow:Lee_SpikeFlowNet_ECCV_2020}
Chankyu Lee, Adarsh~Kumar Kosta, Alex~Zihao Zhu, Kenneth Chaney, Kostas
  Daniilidis, and Kaushik Roy.
\newblock Spike-flownet: event-based optical flow estimation with
  energy-efficient hybrid neural networks.
\newblock In {\em European Conference on Computer Vision (ECCV)}, pages
  366--382, 2020.

\bibitem{eventflow:Bardow_simultaneousflowintensity_2016}
Patrick Bardow, Andrew~J Davison, and Stefan Leutenegger.
\newblock Simultaneous optical flow and intensity estimation from an event
  camera.
\newblock In {\em IEEE Conference on Computer Vision and Pattern Recognition
  (CVPR)}, pages 884--892, 2016.

\bibitem{flow:poggi_sensorflow_ICCV_2021}
Matteo Poggi, Filippo Aleotti, and Stefano Mattoccia.
\newblock Sensor-guided optical flow.
\newblock In {\em IEEE International Conference on Computer Vision (ICCV)},
  pages 7908--7918, 2021.

\bibitem{flow:conti_sensorguide_Lidar_IROS_2022}
Andrea Conti, Matteo Poggi, Filippo Aleotti, and Stefano Mattoccia.
\newblock Unsupervised confidence for lidar depth maps and applications.
\newblock In {\em IEEE/RJS International Conference on Intelligent Robots and
  Systems (IROS)}, 2022.

\bibitem{multimodal_survey_2022}
Peng Xu, Xiatian Zhu, and David~A Clifton.
\newblock Multimodal learning with transformers: {A} survey.
\newblock {\em arXiv preprint arXiv:2206.06488}, 2022.

\bibitem{hori2017attention}
Chiori Hori, Takaaki Hori, Teng-Yok Lee, Ziming Zhang, Bret Harsham, John~R
  Hershey, Tim~K Marks, and Kazuhiko Sumi.
\newblock Attention-based multimodal fusion for video description.
\newblock In {\em IEEE International Conference on Computer Vision (ICCV)},
  pages 4193--4202, 2017.

\bibitem{wei2020multi}
Xi~Wei, Tianzhu Zhang, Yan Li, Yongdong Zhang, and Feng Wu.
\newblock Multi-modality cross attention network for image and sentence
  matching.
\newblock In {\em IEEE Conference on Computer Vision and Pattern Recognition
  (CVPR)}, pages 10941--10950, 2020.

\bibitem{eventapp:sun_eventdeblurattention_eccv_2022}
Lei Sun, Christos Sakaridis, Jingyun Liang, Qi~Jiang, Kailun Yang, Peng Sun,
  Yaozu Ye, Kaiwei Wang, and Luc~Van Gool.
\newblock Event-based fusion for motion deblurring with cross-modal attention.
\newblock In {\em European Conference on Computer Vision (ECCV)}, pages
  412--428, 2022.

\bibitem{representation_learning_yoshua}
Yoshua Bengio, Aaron~C. Courville, and Pascal Vincent.
\newblock Representation learning: {A} review and new perspectives.
\newblock {\em IEEE Transactions on Pattern Analysis and Machine Intelligence
  (TPAMI)}, 35(8):1798--1828, 2013.

\bibitem{hu2017learning}
Weihua Hu, Takeru Miyato, Seiya Tokui, Eiichi Matsumoto, and Masashi Sugiyama.
\newblock Learning discrete representations via information maximizing
  self-augmented training.
\newblock In {\em International Conference on Machine Learning (ICML)}, pages
  1558--1567, 2017.

\bibitem{mine_mutual_information}
Mohamed~Ishmael Belghazi, Aristide Baratin, Sai Rajeshwar, Sherjil Ozair,
  Yoshua Bengio, Aaron Courville, and Devon Hjelm.
\newblock Mutual information neural estimation.
\newblock In {\em International Conference on Machine Learning (ICML)}, pages
  531--540, 2018.

\bibitem{fusion:zhang_rgbdsali_ICCV_2021}
Jing Zhang, Deng-Ping Fan, Yuchao Dai, Xin Yu, Yiran Zhong, Nick Barnes, and
  Ling Shao.
\newblock Rgb-d saliency detection via cascaded mutual information
  minimization.
\newblock In {\em IEEE International Conference on Computer Vision (ICCV)},
  pages 4338--4347, 2021.

\bibitem{vaswani_attention_NeurIPS_2017}
Ashish Vaswani, Noam Shazeer, Niki Parmar, Jakob Uszkoreit, Llion Jones,
  Aidan~N Gomez, {\L}ukasz Kaiser, and Illia Polosukhin.
\newblock Attention is all you need.
\newblock {\em Advances in Neural Information Processing Systems (NeurIPS)},
  30, 2017.

\bibitem{dosovitskiy_imagetransformer_iclr_2020}
Alexey Dosovitskiy, Lucas Beyer, Alexander Kolesnikov, Dirk Weissenborn,
  Xiaohua Zhai, Thomas Unterthiner, Mostafa Dehghani, Matthias Minderer, Georg
  Heigold, Sylvain Gelly, et~al.
\newblock An image is worth 16x16 words: Transformers for image recognition at
  scale.
\newblock In {\em International Conference on Learning Representations (ICLR)},
  2021.

\bibitem{hjelm2018learning}
R~Devon Hjelm, Alex Fedorov, Samuel Lavoie-Marchildon, Karan Grewal, Phil
  Bachman, Adam Trischler, and Yoshua Bengio.
\newblock Learning deep representations by mutual information estimation and
  maximization.
\newblock In {\em International Conference on Learning Representations (ICLR)},
  2019.

\bibitem{ba_mmm_nips03}
David Barber and Felix~V. Agakov.
\newblock The im algorithm: A variational approach to information maximization.
\newblock In {\em Advances in Neural Information Processing Systems (NeurIPS)},
  pages 201--208, 2003.

\bibitem{cheng2020club}
Pengyu Cheng, Weituo Hao, Shuyang Dai, Jiachang Liu, Zhe Gan, and Lawrence
  Carin.
\newblock Club: A contrastive log-ratio upper bound of mutual information.
\newblock In {\em International Conference on Machine Learning (ICML)}, pages
  1779--1788, 2020.

\bibitem{ba2016layer}
Jimmy~Lei Ba, Jamie~Ryan Kiros, and Geoffrey~E Hinton.
\newblock Layer normalization.
\newblock {\em arXiv preprint arXiv:1607.06450}, 2016.

\bibitem{zamir_restormer_CVPR_2022}
Syed~Waqas Zamir, Aditya Arora, Salman Khan, Munawar Hayat, Fahad~Shahbaz Khan,
  and Ming-Hsuan Yang.
\newblock Restormer: Efficient transformer for high-resolution image
  restoration.
\newblock In {\em IEEE Conference on Computer Vision and Pattern Recognition
  (CVPR)}, pages 5728--5739, 2022.

\bibitem{burda2015importance}
Yuri Burda, Roger Grosse, and Ruslan Salakhutdinov.
\newblock Importance weighted autoencoders.
\newblock In {\em International Conference on Learning Representations (ICLR)},
  2016.

\bibitem{alemi2017deep}
Alexander~A. Alemi, Ian Fischer, Joshua~V. Dillon, and Kevin Murphy.
\newblock Deep variational information bottleneck.
\newblock In {\em International Conference on Learning Representations (ICLR)},
  2017.

\bibitem{VAE1}
Diederik Kingma and Max Welling.
\newblock Auto-encoding variational bayes.
\newblock In {\em International Conference on Learning Representations (ICLR)},
  2014.

\bibitem{interaction_information}
R.W. Yeung.
\newblock A new outlook on shannon's information measures.
\newblock {\em IEEE Trans. on Information Theory}, 37(3):466--474, 1991.

\bibitem{liu2019meteornet}
Xingyu Liu, Mengyuan Yan, and Jeannette Bohg.
\newblock Meteornet: Deep learning on dynamic 3d point cloud sequences.
\newblock In {\em IEEE Conference on Computer Vision and Pattern Recognition
  (CVPR)}, pages 9246--9255, 2019.

\bibitem{scene:liu_flownet3d_cvpr_2019}
Xingyu Liu, Charles~R Qi, and Leonidas~J Guibas.
\newblock Flownet3d: Learning scene flow in 3d point clouds.
\newblock In {\em IEEE Conference on Computer Vision and Pattern Recognition
  (CVPR)}, pages 529--537, 2019.

\bibitem{eventdatasets:Gehrig_VideoToEvent_CVPR_2020}
Daniel Gehrig, Mathias Gehrig, Javier Hidalgo-Carri{\'o}, and Davide
  Scaramuzza.
\newblock Video to events: Recycling video datasets for event cameras.
\newblock In {\em IEEE Conference on Computer Vision and Pattern Recognition
  (CVPR)}, pages 3586--3595, 2020.

\bibitem{greff2022kubric}
Klaus Greff, Francois Belletti, Lucas Beyer, Carl Doersch, Yilun Du, Daniel
  Duckworth, David~J Fleet, Dan Gnanapragasam, Florian Golemo, Charles
  Herrmann, et~al.
\newblock Kubric: A scalable dataset generator.
\newblock In {\em IEEE Conference on Computer Vision and Pattern Recognition
  (CVPR)}, pages 3749--3761, 2022.

\bibitem{shen2021cfnet}
Zhelun Shen, Yuchao Dai, and Zhibo Rao.
\newblock Cfnet: Cascade and fused cost volume for robust stereo matching.
\newblock In {\em IEEE Conference on Computer Vision and Pattern Recognition
  (CVPR)}, pages 13906--13915, 2021.

\end{thebibliography}
}

\end{document}